\def\IEEE{1} 

\if\IEEE1
    \documentclass[letterpaper, 10pt, conference]{ieeeconf}  
    \IEEEoverridecommandlockouts                              

\else
    \documentclass[a4paper, fleqn]{cas-dc}
\fi

\if\IEEE0
\usepackage[numbers]{natbib}
\fi

\usepackage{lipsum}  
\usepackage{amsmath,amssymb,amsfonts}
\usepackage{amssymb}
\usepackage{algorithmic}
\usepackage{graphicx}
\usepackage{textcomp}
\usepackage{color}
\usepackage[dvipsnames]{xcolor}
\usepackage{url}
\usepackage{bm}
\usepackage{hyperref}
\usepackage{acronym}
\if\IEEE1
    \usepackage{easyReview}
\fi
\usepackage{multirow,multicol}
\usepackage{tabularx}
\usepackage[american, europeanresistors, straightvoltages]{circuitikz}
\usepackage{siunitx}
\usepackage{subcaption}
\hypersetup{
    pdftitle={bc-nmpc},}

\DeclareMathOperator*{\argmin}{arg\,min}

\usepackage{todonotes}

\newcommand{\Rthree}[0]{\mathbb{R}^3}

\newcommand{\RTwo}[1]{{#1}}
\newcommand{\RThree}[1]{{#1}}
\newcommand{\RCommon}[1]{{{#1}}}

\usepackage{tikz}
\usepackage{pgfplots}
\pgfplotsset{compat=1.14}
\usetikzlibrary{backgrounds,arrows,automata,shapes,positioning,calc,through,spy,shapes,shapes.geometric,shapes.multipart,fit,patterns,fadings}
\pgfdeclarelayer{background}
\pgfdeclarelayer{foreground}
\pgfsetlayers{background, main, foreground}

\if\IEEE1
\title{\LARGE \bf
BC-NMPC: Battery-Constrained NMPC with Propulsion Prediction and Replanning for High-Speed Flight}
\author{
   Parakh M. Gupta$^{\ast}$,
   Matěj Mihulka,
   Matej Novosad,
   Robert Pěnička,
   Martin Saska
    \thanks{  
   $^{\ast}$Corresponding author: guptapar@fel.cvut.cz.
     The authors are with the Multi-robot Systems Group, Czech Technical University in Prague, Czech Republic (\protect\url{https://mrs.fel.cvut.cz}).
    This work was funded by the European Union under the project Robotics and advanced industrial production (reg. no. CZ.02.01.01/00/22\_008/0004590), and by the Czech Science Foundation (GAČR) under research project no. 23-06162M.
    We would also like to thank our team member Tomáš Kestřánek for designing the hardware for the identification of the propulsion parameters.
     }
}
\else
    \title[mode = title]{BC-NMPC: Battery-Constrained NMPC with Propulsion Prediction and Replanning for High-Speed Flight}
    \tnotemark[1]
    \tnotetext[1]{This work was funded by the European Union under the project Robotics and advanced industrial production (reg. no. CZ.02.01.01/00/22\_008/0004590), and by the Czech Science Foundation (GAČR) under research project no. 23-06162M.
    We would also like to thank our team member Tomáš Kestřánek for designing the hardware for the identification of the propulsion parameters.}

    \author[1]{Parakh M. Gupta}[orcid=0000-0002-6481-2281]
    \fnmark[1]
    \ead{guptapar@fel.cvut.cz}
    \ead[url]{https://mrs.fel.cvut.cz}

    \affiliation[1]{organization={Multi-Robot Systems Group, Czech Technical University},
                    city={Prague},
                    postcode={120 00}, 
                    country={Czech Republic}}

    \author[1]{Matěj Mihulka}

    \author[1]{Matej Novosad}

    \author[1]{Robert Pěnička}
    \author[1]{Martin Saska}

    \fntext[fn1]{Corresponding author: guptapar@fel.cvut.cz}

\fi

\begin{document}


\maketitle
\thispagestyle{empty}
\pagestyle{empty}

\newcommand{\position}{p}
\newcommand{\positionvector}{\bm{\position}}
\newcommand{\quaternion}{q}
\newcommand{\orientation}{\bm{\quaternion}}
\newcommand{\linearvel}{v}
\newcommand{\linearvelvector}{\bm{\linearvel}}
\newcommand{\angularvel}{\omega}
\newcommand{\angularvelvector}{\bm{\angularvel}}
\newcommand{\integralerror}{z}
\newcommand{\integralerrorvector}{\bm{\integralerror}}
\newcommand{\statederivative}{\dot{\statestd}}
\newcommand{\statestd}{\bm{x}}
\newcommand{\uavstate}{\statestd}
\newcommand{\statepid}{\bm{x}}
\newcommand{\commandedangularvel}{\angularvel_c}
\newcommand{\propvel}{\Omega}
\newcommand{\propvelvector}{\bm{\propvel}}
\newcommand{\propvelcmd}{\propvel_c}
\newcommand{\propvelcmdvector}{\bm{\propvelcmd}}
\newcommand{\propdragcoef}{C_q}
\newcommand{\dragforcevector}{\bm{f}_{D}}
\newcommand{\dragtorque}{\tau_d}
\newcommand{\commandedtorquevector}{\bm{\tau_c}}
\newcommand{\torquevector}{\bm{\tau}}
\newcommand{\uavinput}{\bm{u}}
\newcommand{\motorforce}{f}
\newcommand{\thrustvector}{\bm{f}}
\newcommand{\commandedthrustvector}{\bm{f_c}}
\newcommand{\normalizedrpm}{r}
\newcommand{\normalizedrpmvector}{\bm{\normalizedrpm}}
\newcommand{\throttle}{t_h}
\newcommand{\throttlevec}{\bm{\throttle}}
\newcommand{\commandedcollectivethrottle}{\throttle_c}
\newcommand{\collectivethrust}{T}
\newcommand{\collectivethrustmax}{\collectivethrust_\text{max}}
\newcommand{\commandedcollectivethrust}{\collectivethrust_c}
\newcommand{\bodythrustvector}{\bm{f}_{T}}
\newcommand{\motorallocationmat}{\bm{M}}
\newcommand{\mixermat}{\bm{G}}
\newcommand{\inertiamat}{\bm{J}}
\newcommand{\motorpropinertiamat}{J_{mp}}
\newcommand{\mpchorizon}{N}
\newcommand{\errpenmat}{\mathbf{Q}}
\newcommand{\inputpenmat}{\mathbf{R}}
\newcommand{\terminalpenmat}{\mathbf{T}}
\newcommand{\errormat}{\mathbf{\Tilde{\statestd}}}
\newcommand{\errorinputmat}{\mathbf{\Tilde{\uavinput}}}
\newcommand{\statestddesired}{\overset{*}{\statestd}}
\newcommand{\uavinputdesired}{\overset{*}{\uavinput}}
\newcommand{\taumot}{k_\text{mot}}
\newcommand{\thrustcoef}{c_f}
\newcommand{\piderror}{e}
\newcommand{\piderrorvec}{\bm{\piderror}}

\newcommand{\batterytempcoef}{K_t}
\newcommand{\batterysoc}{s_\text{oc}}
\newcommand{\estbatterysoc}{\hat{s}_\text{oc}}
\newcommand{\batteryvoltage}{v_\text{oc}}
\newcommand{\estbatteryvoltage}{\hat{v}_\text{oc}}
\newcommand{\circuitcurrent}{i}
\newcommand{\meascircuitcurrent}{\overset{*}{i}}
\newcommand{\estcircuitcurrent}{\hat{i}}
\newcommand{\circuitvoltage}{v_\text{cc}}
\newcommand{\estcircuitvoltage}{\hat{v}_\text{cc}}
\newcommand{\meascircuitvoltage}{\overset{*}{v}_\text{cc}}
\newcommand{\batteryresistance}{r_b}
\newcommand{\measbatteryresistance}{\overset{*}{r}_b}
\newcommand{\estbatteryresistance}{\hat{r}_b}
\newcommand{\motorresistance}{r_m}
\newcommand{\motorvoltage}{\circuitvoltage}
\newcommand{\motorconstant}{k_m}
\newcommand{\motortorque}{\tau_m}
\newcommand{\motorzerocurrent}{i_0}
\newcommand{\powerin}{p_\text{in}}

\newcommand{\rpmtoforcecoef}{f}
\newcommand{\thrusttopowercoef}{p}
\newcommand{\motorresistancecoef}{r}
\newcommand{\volthrotcoef}{t}
\newcommand{\voltthrustcoef}{f}

\newcommand{\volttosoccoef}{s}
\newcommand{\soctovoltcoef}{q}
\newcommand{\soctorescoef}{d}

\newcommand{\coeffthv}{173.64}
\newcommand{\interceptthv}{673.59}

\newcommand{\powerinsingle}{p_{\text{in},i}}
\newcommand{\motorforcesingle}{f_i}
\newcommand{\accellimit}{a^T_{\text{max}}}
\newcommand{\mass}{m}
\newcommand{\accelgrav}{g}
\newcommand{\weightmatrix}{W}
\newcommand{\terminalweightmatrix}{W_e}
\newcommand{\uavoutput}{\bm{y}}

\acrodef{pid}[PID]{Proportional Integral Derivative}
\acrodef{uav}[UAV]{Uncrewed Aerial Vehicle}
\acrodef{vtol}[VTOL]{Vertical Take-Off and Landing}
\acrodef{mpc}[MPC]{Model Predictive Controller}
\acrodef{nmpc}[NMPC]{Nonlinear Model Predictive Controller}
\acrodef{dfbc}[DFBC]{Differential-Flatness-Based Controller}
\acrodef{indi}[INDI]{Incremental Nonlinear Dynamic Inversion}
\acrodef{ocp}[OCP]{Optimal Control Problem}
\acrodef{imu}[IMU]{Inertial Measurement Unit}
\acrodef{rti}[RTI]{Real-Time Iteration}
\acrodef{rpm}[RPM]{Rotations Per Minute}
\acrodef{rmse}[RMSE]{Root Mean Square Error}
\acrodef{qp}[QP]{Quadratic Programming}
\acrodef{gnss}[GNSS]{Global Navigation Satellite System}
\acrodef{mav}[MAV]{Micro Aerial Vehicle}
\acrodef{a2rl}[A2RL]{Abu Dhabi Autonomous Racing League}
\acrodef{rl}[RL]{Reinforcement Learning}
\acrodef{esc}[ESC]{Electronic Speed Controller}
\acrodef{ir}[IR]{Internal Resistance}
\acrodef{soc}[SOC]{State of Charge}
\acrodef{irt}[IR]{Internal Resistance}
\acrodef{mae}[MAE]{Mean Absolute Error}
\acrodef{rmse}[RMSE]{Root Mean Square Error}
\acrodef{bem}[BEM]{Blade Element Momentum}
\acrodef{bldc}[BLDC]{Brushless Direct Current}
\acrodef{fcu}[FCU]{Flight Control Unit}
\acrodef{nimh}[NiMH]{Nickel-Metal Hydride}
\acrodef{nicd}[NiCd]{Nickel-Cadmium}
\acrodef{lithiumion}[Li-Ion]{Lithium-Ion}
\acrodef{lipo}[LiPo]{Lithium-Polymer}
\acrodef{LiFePO4}[LiFePO4]{Lithium Iron Phosphate}
\acrodef{rtt}[RTT]{Required Thrust Threshold}

\begin{abstract}
\RCommon{
Trajectory tracking performance of \acp{uav} degrades during an agile high-speed flight due to the depletion of the battery and subsequent loss of maximum available thrust.
In applications such as drone racing, this leads to a failure to complete the race due to possible collisions with obstacles.
In this paper, we present a novel method for integrating battery and propulsion system models into a \ac{nmpc} framework to enable real-time prediction of the voltage, current, power, and maximum available thrust of the platform. 
Our proposed approach achieves lower trajectory tracking error as a result of its real-time thrust awareness, and with the help of trajectory replanning, it allows the \ac{uav} to fly in time-optimal regime throughout the mission.
The accuracy of this proposed model was verified in real-world flight experiments, while the effectiveness of the replanning algorithm was evaluated in simulation.
By the end of the battery capacity, compared to an unaware controller, our novel controller achieved a \SI{25}{\percent} reduction in mean position error without replanning, and an \SI{88}{\percent} reduction with replanning.
}
\end{abstract}

\section{Introduction}

\label{sec:introduction}

Multi-rotor \acfp{uav} have become ubiquitous in the modern world for applications such as oil-spill and fire-monitoring~\cite{Kingston_2008_perimeter_surveillance}, search-and-rescue~\cite{Rudol_2008_sar}, radiation leak identification~\cite{Han2014, Stibinger_2020_radiation_localization}, infrastructure inspection~\cite{Burri_2012_plant_inspection,Merz_2011_inspection}, aerial surveying~\cite{datsko2024fastCoverage}, and drone racing~\cite{verraest2025skydreamer}.
Commercial advances in computing and control algorithm efficiency, over the last two decades, have enabled onboard computation of autonomous control and autonomous planning to unlock more complex and dynamic applications for \acp{uav}.
Through the medium of various racing competitions in the last 5 years, researchers have shown that they can match and beat the most skilled human racing pilots in equal machinery \cite{verraest2025skydreamer,bahnam2026monorace}.
\begin{figure}
    \centering
    \includegraphics[width=0.5\textwidth]{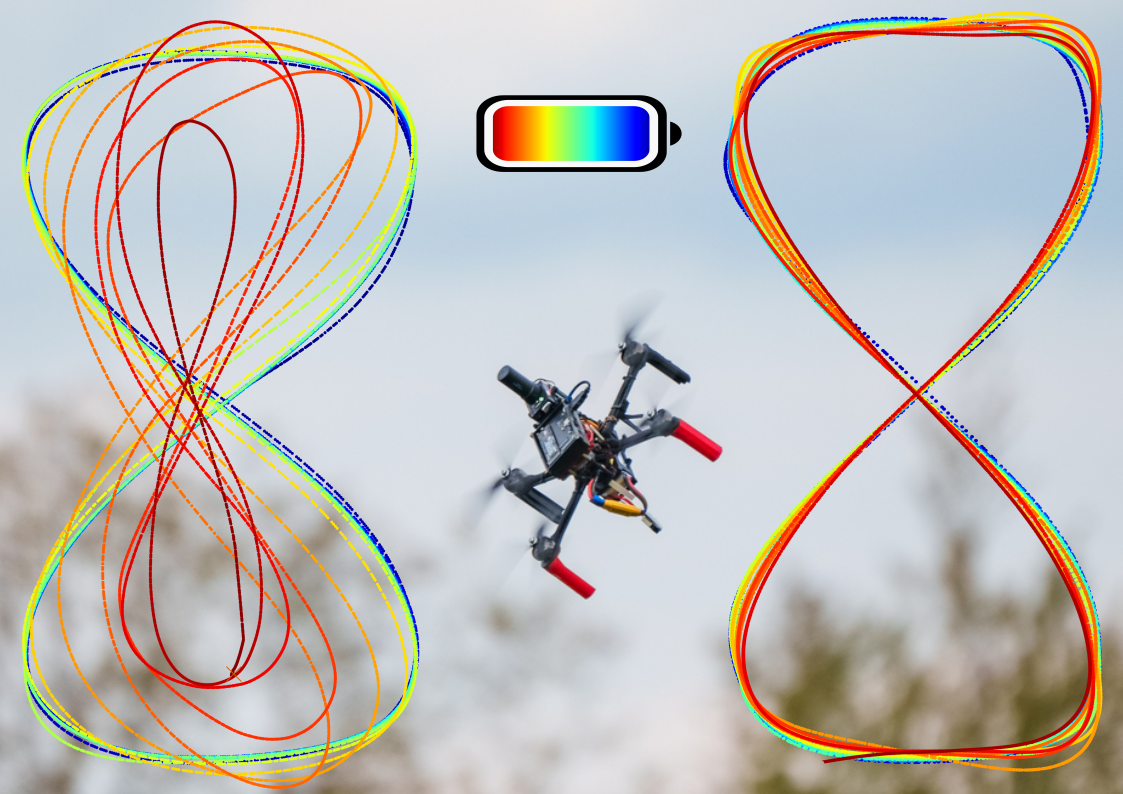}
    \caption{Our custom-built racing aircraft captured during its agile flight experiments in the outdoor, alongside the trajectory overlay of its flight (not to scale). Trajectory without replanning on the left, and trajectory with battery-constrained replanning on the right.}
    \label{fig:intro_pic}
\end{figure}
Flying autonomously at their physical limits, these competition-class \acp{uav} clocked-in speeds of up to \SI{100}{\kilo\meter\per\hour} and accelerations of up to \SI{7}{g}~\cite{verraest2025skydreamer,bahnam2026monorace}.

Most competitors in these competitions employed a \acf{nmpc}, an optimisation-based controller that has gained popularity in real-world agile high-speed flight.
In an \ac{nmpc}, control actions are calculated (usually at \SI{100}{\hertz}) using predictions from physics models by finding desired system states that minimise the trajectory tracking error in the future \cite{Sun2022NMPCcomparativeStudy}.
The fidelity and complexity of the dynamic model are a critical factor in the performance of an \ac{nmpc}, but the strict requirement of real-time on-board computability hinders complex models from being deployed in outdoor robotics.
Therefore, some assumptions are usually made to simplify the models, and one of these assumptions is the focus of this paper. 
The current approaches in the literature use offline electro-mechanical models to map and look-up parameters for the entire propulsion system and therefore, they employ stationary limits on the linear and angular accelerations of the platform \cite{agilicious, cpc}.
However, the rate of depletion of the battery affects the maximum thrust available per motor, which, in-turn, affects the maximum linear and angular acceleration that the vehicle can generate.
Usually, these limits shrink during the flight, and assumption of static limits can render a pre-planned high-speed trajectory infeasible halfway through the flight.
For example, if a pre-planned trajectory manoeuvres around an obstacle, the loss of maximum available torque can lead to a shallower turn and a collision with the obstacle.

To model the propulsion system, recent research separates the problem into the electrical \cite{Battery_modelling} and the aerodynamical \cite{bauersfeldNeuroBEM2021} subsystems of the propulsion system.
As part of the electrical subsystem, battery models have been proposed with varying degrees of complexity to capture the behaviour of the battery \cite{Battery_modelling}.
These models cover chemistries such as \ac{nimh}, \ac{nicd}, \ac{lipo}, and \ac{lithiumion} for low to medium discharge rates, but, either the complexity of these models or their constant-discharge assumption prevents them from being used in real-time on-board the \acp{uav}.
Many of these models also contain transient dynamics that require discrete-time implementation with very small time-steps that might not be feasible in real-time control \cite{Doerffel2006, galushkin2014generalized, Battery_circuits, Dees2002}.

However, \acp{uav} are a specific application where rapidly-varying high-discharge rates are common, and the literature is rather thin for this specific discharge regime.
Traditionally, \acp{uav} use \ac{lipo} batteries due to their high-discharge rates, and this battery chemistry exhibits a larger voltage drop in its usable capacity range.
Lithium Iron Phosphate (LiFePO4) offers a more stable voltage (no load) across capacity range with better stability and resistance in overcharge and high-temperature conditions, but they offer lower energy density in comparison to LiPo batteries \cite{SERGI2016235}.
\acf{lithiumion}, on the other hand, offers higher energy density than LiPo batteries along with more stable voltage (no load) across capacity range.
Due to these properties, LiIon are preferred battery types for long-distance \ac{uav} missions, but they offer significantly lower discharge rate limits as their voltage drop is much higher under load, and so, the \ac{uav} thrust limits are more sensitive to this battery type.

Conversely, aerodynamic and mechanical properties of the propulsion system have been studied in great depth in the context of \acp{uav}.
Using neural networks and \ac{bem} theory, Bauersfeld et al. \cite{bauersfeldNeuroBEM2021} demonstrated high-accuracy models that can predict the performance of the propulsion system to achieve an improvement in trajectory tracking accuracy. 
This model can be evaluated relatively quickly onboard the \ac{uav} but is infeasible for deployment in an \ac{nmpc} as multiple evaluations required for each iteration of the optimisation problem are computation-heavy.
The same researchers built on their work to include battery modelling for estimating the flight range and endurance of the aircraft \cite{bauersfeld2022range}.
However, their approach ran offline and was demonstrated for either hover or constant medium-speed flight conditions ($\approx$ \SI{18}{\meter\per\second}).
Another often-used solution to this problem is to handle the obstacle avoidance in the trajectory planning, but then trajectory tracking errors need to be prioritised over performance, since any deviation in an obstacle-rich environment can lead to collision. 
Also, depletion of thrust leads to depletion of acceleration and torques which can lead to lag and trajectory error build-up, and therefore cause collisions when the controller attempts to take a shortcut to sacrifice tracking error for lag error minimisation.
This necessitates the integration of propulsion-limit awareness into the controller, and ultimately, the real roadblock for the integration of predictive battery modelling into the \ac{nmpc} framework is the trade-off between the accuracy of the model and its computational complexity.

To solve this, we present a novel modelling approach for the electro-mechanical characteristics of the propulsion system which
\begin{itemize}
    \item is computationally feasible through the use of multi-variate polynomials,
    \item isolates the electrical and mechanical systems for modular and easier identification of the system parameters,
    \item allows effective predictions of electrical and mechanical aspects of flight in real-time,
    \item permits integration into an \ac{nmpc} for online re-calculation of collective available thrust, 
    \item \RCommon{and, enables the vehicle to fly consistently at the limits of the thrust throughout the battery capacity range to maintain time-optimality.}
\end{itemize}
Owing to this new model, our \ac{nmpc}-based approach can predict the propulsion constraints of the platform at \SI{100}{\hertz} and stay at the limit of propulsion throughout the flight. 
Additionally, we extend an offline trajectory planner \cite{PMM} to dynamically replan a trajectory in flight and prevent the accumulation of trajectory tracking error as the available thrust of the platform declines throughout the flight.

\section{Related work}\label{sec:related_work}
The field of multi-rotor \acp{uav} combines knowledge and tools from various fields such as electrical engineering, mechanical engineering, aerodynamics, control theory, trajectory planning, perception, and state estimation.
As such, the research relevant to \acp{uav} is often domain-specific, and therefore, the following section describes the state-of-the-art methods on each subsystem individually, and then, evaluates literature on integrated system models.
\subsection{Battery Modelling}
Battery modelling  \cite{Battery_modelling} can be categorised into four main approaches: electrochemical modelling, electrical-circuit modelling, analytical modelling, and stochastic modelling.

Electrochemical modelling can account for chemical and design characteristics such as transport of ions, electrons, electrode geometry, electrolyte concentration, diffusion coefficients, and other kinetic electrochemical processes.
These models provide the highest accuracy but are computationally expensive, and require extensive calibration with sometimes over 50 battery parameters \cite{Battery_modelling, Dees2002}.

The electrical-circuit models require a lot of experimental data and they can be computationally efficient, but they are less accurate than electrochemical models \cite{Battery_circuits}.
The analytical models have been effective and there are three important approaches to consider. 
The first method applies Peukert's law and a review of this method by Doerffel et al. \cite{Doerffel2006} concluded that Peukert's equation could not be used to predict the \ac{soc} of a battery unless the study was conducted at constant current and constant temperature. 
Galushkin et al. \cite{galushkin2014generalized} presented an analytical model to generalise Peukert's law for lithium batteries and extended it to capture the discharge behaviour of Lithium cells across various discharge rates and temperatures.
The second analytical method, proposed by Rakhmatov and Vrudhula \cite{Rakhmatov2001}, used the diffusion process of the active material in the battery. 
It is based on a combination of three differential equations that account for both rate capacity effect and recovery effect during battery discharge.
However, this model remains computationally complex.
The last method is called the Kinetic Battery Model or the two-well method, originally proposed by Manwell \cite{Manwell1993} et al., and it uses chemical kinetic processes. 
It divides the battery charge into two wells: an available-charge well which supplies charge directly to the load, and a bound-charge well, which supplies charge to the first well when certain conditions are met.
While this method is good at predictions, unfortunately, it does not hold for LiIon or LiPo batteries since they have sloped discharge curves \cite{Battery_modelling}.

The stochastic methods also use analytical equations, but unlike purely analytical models, they represent the discharge and recovery effects as stochastic processes. 
As an example, the study by Pozzi and Raimondo \cite{Pozzi2022} proposed a stochastic model predictive control strategy to explicitly account for parameter uncertainties using polynomial chaos expansion.
That said, stochastic models are very limited in their application/scope, and do not tackle varying discharge currents seen in real-world applications.

For the purpose of battery estimation in this presented work, since we mainly focus on estimating the open-circuit voltage and internal resistance of the battery during flight, a combination of analytical modelling and electrical-circuit modelling offers the right tradeoff for balancing computational complexity and accuracy.
Rahmoun et al. \cite{Battery_circuits} described three common types of electrical-circuit models namely the \ac{ir} model, the one time-constant model, and the two time-constant model.
However, since both the estimation of open-circuit voltage and internal resistance rely on estimation of \ac{soc}, \ac{ir} models can suffer heavily if \ac{soc} of the battery is estimated poorly. 
In a traditional closed-loop system, the \ac{soc} is estimated using coulomb counting, which can accumulate errors over time and over multiple charge-discharge cycles.
This issue can be eliminated for the \acp{uav} as a different battery is replugged before every flight, and the \ac{soc} can be reset at the beginning of each flight after a single discharge cycle.
As such, the simplicity of the \ac{ir} model is leveraged in this work as a starting point for a scientific inquiry into real-time battery predictions onboard an aircraft.

\subsection{Motor-propeller modelling}
Motors and Propellers are often modelled as a single sub-system when quantifying the mechanical output of a propulsion system.
However, it is the motor that performs the electrical-to-mechanical energy conversion, while the propeller converts mechanical energy into thrust.
As such, the electro-mechanical characteristics of the motor are the bridge to understanding the performance of the entire propulsion pipeline.

Coates et al. \cite{coatesPropulsionSystemmodeling2019} and Moseler et al. \cite{moselerApplicationModelbasedFault2000} proposed a novel method for identifying and predicting electro-mechanical characteristics of a \ac{bldc} motor and provided a comparative analysis against Beard \& McLain's model \cite{Beard2012smallUAVs}.
Their model offers a great baseline for linking electrical and mechanical properties of the entire propulsion system, but struggles to predict current consumption and requires expensive testing equipment to identify a high number of parameters.
Gabriel et al. \cite{gabrielBrushlessDCMotor2011} presented a method for evaluating and modelling the efficiency of a \ac{bldc} motor using a four-constant model but their work did not focus on inter-linking the electrical properties of the motor to its mechanical properties.

For propeller modelling, two separate models viz. Blade Element Model and Momentum Model have been used to capture the aerodynamic performance of propellers.
Blade element theory divides the propeller into infinitesimally thin elements, assuming no interaction between them and independent fluid flow. 
Axial and tangential forces are then calculated for each element, and integrated along the blade span to obtain total thrust and torque.
Momentum theory, on the other hand, considers the propeller as a disc that imparts momentum to the airflow, leading to changes in velocity and pressure across the disc, thus producing thrust.
In a study by Faisal Mahmuddin \cite{BEM}, these methods are combined into \ac{bem} theory to model the aerodynamic performance of propellers more accurately and find the propeller profile suited for wind turbines.
Due to its complexity and the number of parameters that must be identified, this method is mainly used for offline analysis.
However, when the parameters are accurately identified, it can provide a highly representative model of the propeller.

For a fast evaluation of the integrated motor--propeller model, Bauersfeld et al. \cite{bauersfeldNeuroBEM2021} presented a new model using a combination of neural networks and \ac{bem} theory.
While this model performs very well for prediction of aerodynamic and drag properties during high-speed agile flight and can adapt to rapid change in inlet air velocity at the propeller, it does not include the electrical characteristics of the propulsion system.
Amezquita-Brooks et al. \cite{Propulsion} also presented an expanded model of the mechanical side of the combined motor--propeller system where they discussed the expansion of the flight envelope to various horizontal flight speeds and propeller orientations for the purposes of \ac{vtol} aircrafts.
Their work focuses on horizontal flight and does not generalise for multi-rotor \acp{uav}.

In view of this, we propose a simplified polynomial approach to characterising the relationships between the electrical and the mechanical properties of the motor-propeller system.
This approach allows for fast real-time evaluation inside the \ac{nmpc} and can be used for predictions.
The transient properties of the motor-propeller system are not considered in this model as that would require identification in a wind tunnel or a flight volume with a motion capture system.

\subsection{System model}\label{sec:system_model}
As mentioned earlier, battery modelling and motor-propeller modelling have been separate areas of research, however, the problem of estimating power, energy, and flight range of \acp{uav} is a system-level problem that requires modelling of the entire propulsion system as a whole.
For the problem statement of our work, it is essential to note that the available thrust and control authority of a \ac{uav} is influenced by the available instantaneous power, and therefore, instantaneous power needs to be treated differently from energy (available or consumed).
An integrated approach to both power and energy can offer a low-complexity solution for enabling fast real-time computation with reasonable accuracy, while also allowing flight range estimation.

A study by Santos et al. \cite{santos2021energy} presented a similar integrated approach for energy-aware model predictive control with obstacle avoidance.
The study used a simple battery model based on Peukert’s law to devise state-of-charge dynamics which were incorporated into linear MPC. 
This allowed the authors to constrain the state-of-charge, the discharge power, and the change of discharge power. 
Their proposed method minimised the discharge power inside the objective of MPC to stay close to a reference state. 
Unfortunately, this study showed simulation results and did not directly account for change in the thrust with the changing \ac{soc}.

A study by Bauersfeld et al. \cite{bauersfeld2022range} focused on the energy and flight range estimation of multicopters.
The study combined blade-element-momentum theory with motor efficiency and electrical behaviour to create a motor model.
The battery was modelled using a Thevenin equivalent circuit with a one time-constant model.  
These calculations were, however, performed offline, which enabled the use of a more complex battery model. 
Their method aimed to solve offline computations of maximum flight times, and therefore, did not focus on real-time prediction of the electro-mechanical properties of the propulsion system for control purposes.

Another study by Abeywickrama et  al. \cite{Energy_consumption} presented an energy consumption model that incorporated hover, forward flight, acceleration, wind, payload, and ageing effects in order to enable offline planning of energy-efficient missions by minimising high-consumption actions.

While these studies demonstrate either accurate battery prediction or accurate motor-propeller model, to the best of our knowledge, our paper presents the first approach that incorporates battery and motor--propeller models directly into the \ac{nmpc} computation for real-time use during the flight.

\section{Electro-Mechanical Propulsion Model}

\subsection{Electrical model of \ac{uav} propulsion}
The electrical model of the \ac{uav}'s propulsion (see Figure \ref{fig:drone_circuit}) consists of the battery equivalent \ac{ir} circuit, represented by open circuit voltage $\batteryvoltage$ and the internal resistance $\batteryresistance$, and motors are represented by four variable resistances $\motorresistance^i$ in a parallel network.
In comparison to the peak power consumption of \SI{800}{\watt} per motor (\SI{50}{\ampere} at \SI{16.0}{\volt}), the power lost in transmission and the power consumed by the onboard computer are negligible, and so, they are not included in the circuit model.
The motor is modelled as a resistive load since the transient characteristics of the inductance modelling are much faster than the time step of the \ac{nmpc} controller, and therefore, they are neglected to maintain real-time computability.
The in-flight measured quantities include the terminal voltage $\motorvoltage$ and the current drawn from the battery $\circuitcurrent$.
The mechanical properties of the propulsion system (motor-propeller) are linked to the electrical properties of the propulsion system (battery) through the inherent relationships between input voltage and output thrust.
Electrically, this system is governed by the Kirchhoff's circuit laws such that

\vspace{-0.5em}
\begin{align}
    \label{eq:volt_div}
    \begin{aligned}
        \frac{1}{\motorresistance} &= \frac{1}{\motorresistance^1} + \frac{1}{\motorresistance^2} + \frac{1}{\motorresistance^3} +  \frac{1}{\motorresistance^4} \\
    \motorvoltage &= \circuitcurrent \motorresistance, \\
    \batteryvoltage &= \circuitvoltage + \circuitcurrent\batteryresistance\text{, and, } \\
    \powerin &= \motorvoltage\circuitcurrent, \\
    \end{aligned}
\end{align}
\vspace{-0.5em}

where $\motorresistance$ is the equivalent parallel resistance of the motor network.
In the following sections, we discuss the proposed motor model and battery model for our \ac{nmpc} formulation.

\begin{figure}
	\centering
    \begin{circuitikz}
    \draw
    (0,2) to[battery, l_=$\batteryvoltage$] (0,0)
    (0,2) to[european resistor, R=$\batteryresistance$] (2,2)
    node at (-0.3,1.5) {+}
    node at (-0.3,0.5) {--}
    (2.00,2) to[short, i=$\circuitcurrent$] (3.5,2)

    (2,2) -- (3.5,2)
    to[vR] (3.5,0)
    -- (2,0)
    node at (3.9,1.75) {$\motorresistance^1$}

    (2,2) -- (4.5,2)
    to[vR] (4.5,0)
    -- (2,0)
    node at (4.9,1.75) {$\motorresistance^2$}

    (2,2) -- (5.5,2)
    to[vR] (5.5,0)
    -- (2,0)
    node at (5.9,1.75) {$\motorresistance^3$}

    (2,2) -- (6.5,2)
    to[vR] (6.5,0)
    -- (2,0)
    node at (6.9,1.75) {$\motorresistance^4$}

    (2,0) -- (0,0)
    (2.25,0) node[circ]{} node[rground]{}

    (2.25,2) -- (2.25,1.75) node[ocirc]{}   
    (2.25,0) -- (2.25,0.25) node[ocirc]{}   
    node at (2.25,1) {$\motorvoltage$}           
    node at (2.25,1.5) {+}
    node at (2.25,0.5) {--};
    \draw[dotted, thick] (-1.0,-0.4) rectangle (1.8,2.6)
    node at (0.4,2.8) {Battery};
    \draw[dotted, thick] (3.1,-0.4) rectangle (7.3,2.6)
    node at (4.4,2.8) {Motors};

\end{circuitikz}
    \caption{\ac{uav} circuit consisting of battery, its internal resistance, and four motors represented by variable resistances in parallel.}
	\label{fig:drone_circuit}
\end{figure}

\subsection{Motor model}
\label{sec:motor_model}
\begin{figure*}[h]
	\centering

	\begin{subfigure}[b]{0.48\textwidth}
		\centering
		\includegraphics[width=\linewidth]{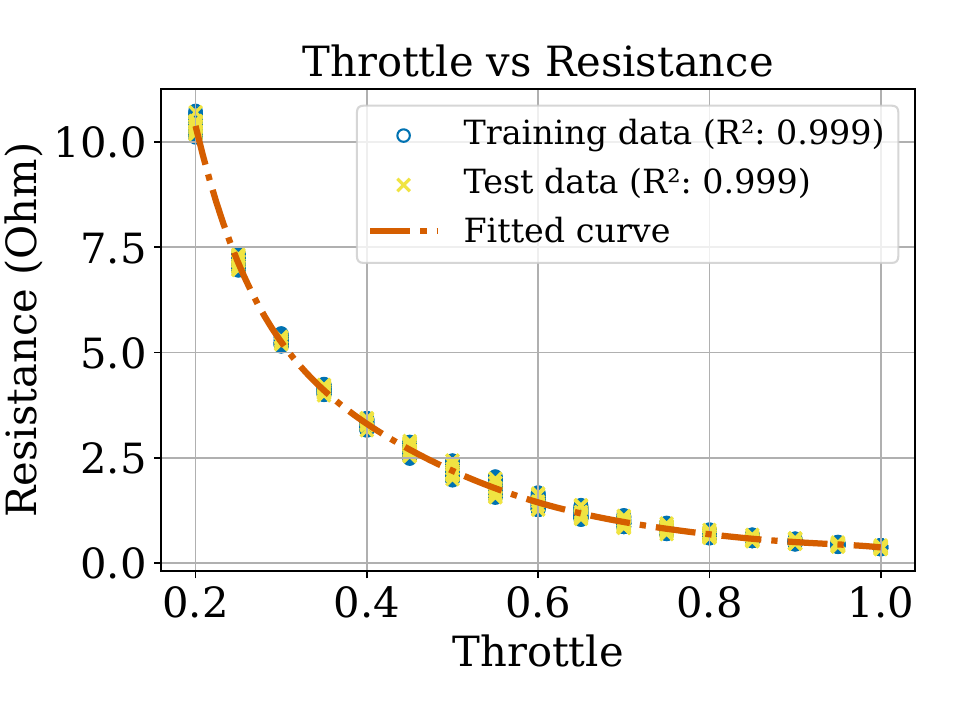}
		\label{fig:plot1}
	\end{subfigure}
	\begin{subfigure}[b]{0.48\textwidth}
		\centering
		\includegraphics[width=\linewidth]{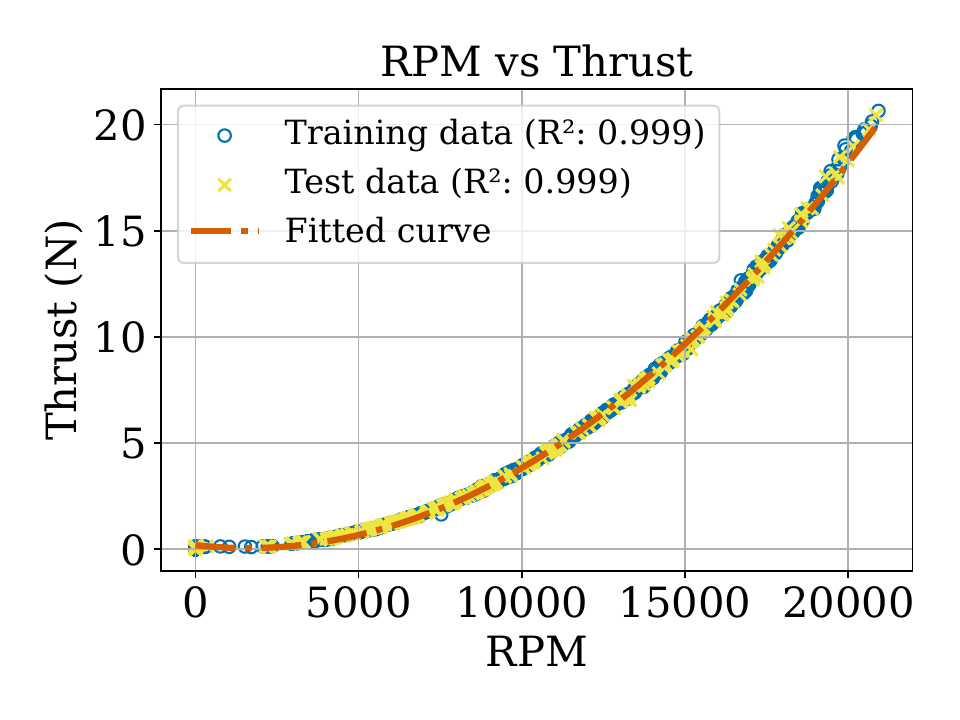}
		\label{fig:plot2}
	\end{subfigure}
	\begin{subfigure}[b]{0.50\textwidth}
		\centering
		\includegraphics[width=\linewidth]{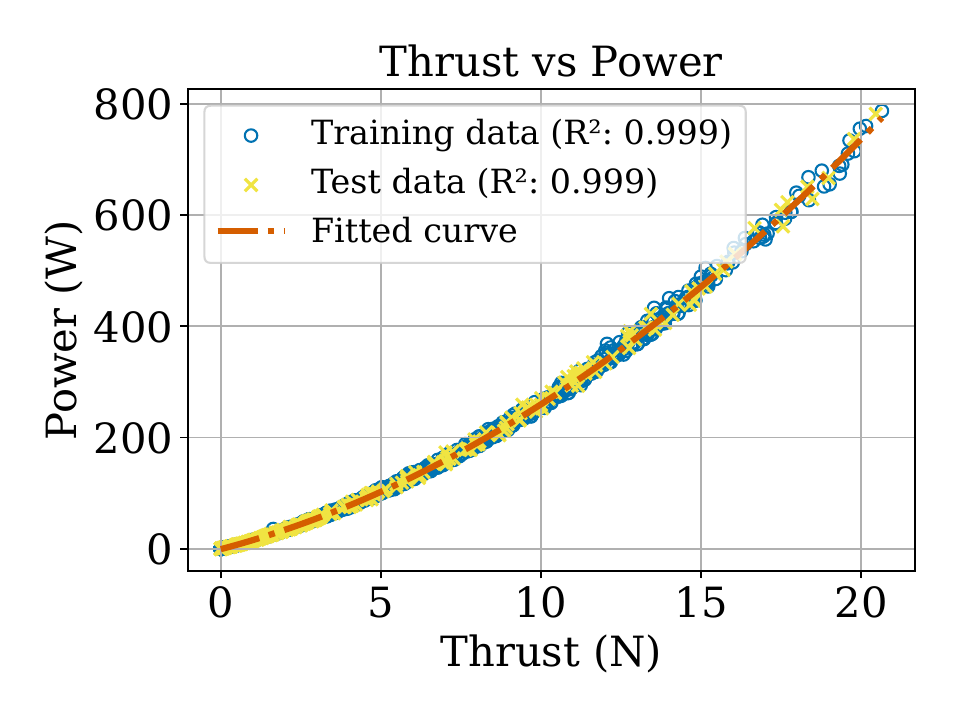}
		\label{fig:plot3}
	\end{subfigure}
	\begin{subfigure}[b]{0.49\textwidth}
		\centering
		\includegraphics[width=1.1\linewidth]{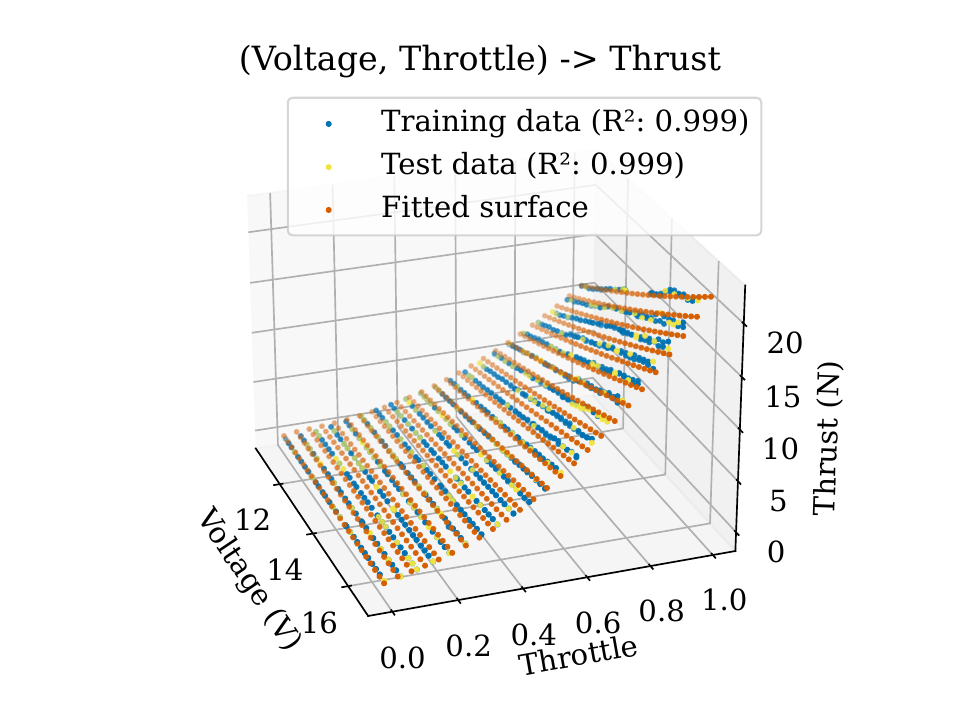}
		\label{fig:plot4}
	\end{subfigure}

	\caption{Fitted curves between throttle, voltage, thrust, resistance, and power obtained from thrust testing showing a close-fit for all assumed relationships.}
	\label{fig:motor_plots}
\end{figure*}
\begin{figure}
    \centering
    \includegraphics[width=0.3\textwidth, angle=0]{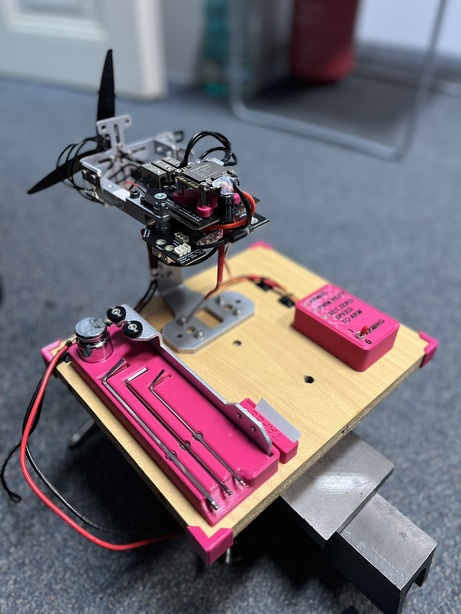}
    \caption{Thrust testing setup used to obtain the motor-propeller characteristics.}
    \label{fig:thrust_tester}
\end{figure}

In this work, each motor-propeller unit is modelled as a resistive load that converts electrical power to mechanical power.
Four of these resistive loads are connected in parallel and their collective properties are a function of the interaction of each individual unit.
To ascertain the performance of the motor-propeller pair, a static thrust stand was used as shown in Figure \ref{fig:thrust_tester}.
In this setup, the electrical input was quantified using the voltage, current, and power measurements, while mechanical output was quantified using the thrust, the associated \ac{rpm}, and the torque produced.

For our custom-built \ac{uav}, and for this thrust stand test, an Emax ECOII 2807 1700kV motor was paired with a Gemfan 7040 Tri-blade propeller.
The data was collated over 30 experimental runs ranging from \SI{11.0}{\volt} to \SI{16.8}{\volt} at \SI{0.2}{\volt} increments, and for each run, the throttle was varied from \SI{0}{\percent} to \SI{100}{\percent} at \SI{5}{\percent} increments.
Electrical and mechanical measurements were collected 2 seconds after the throttle was changed to ensure that the system had reached a steady state.
\RCommon{For simplification of notation in this section, each of the following quantities only denotes the properties of an individual motor-propeller pair.
The obtained data was split into a 4:1 ratio for fitting/training and testing respectively, and the following polynomial functions were obtained from the fitting data between motor resistance $\motorresistance$, motor force $\motorforce$, throttle control signal $\throttle$, terminal voltage $\motorvoltage$, power consumed by the motor $\powerin$, and  \ac{rpm} $\propvel$:}

\begin{align}
    \motorforce      & = a_\rpmtoforcecoef\propvel^2 + b_\rpmtoforcecoef\propvel + c_\rpmtoforcecoef\text{,}  \label{eq:rpm_to_force}                                                                                           \\
    \powerin         & = a_\thrusttopowercoef\motorforce^2 + b_\thrusttopowercoef\motorforce + c_\thrusttopowercoef\text{,} \label{eq:thrust_to_power}                                                                             \\
	\motorresistance & = a_\motorresistancecoef\throttle^7 + b_\motorresistancecoef\throttle^6 + c_\motorresistancecoef\throttle^5 + d_\motorresistancecoef\throttle^4     \notag                      \\
                     & \hphantom~+~e_\motorresistancecoef\throttle^3 + f_\motorresistancecoef\throttle^2 + g_\motorresistancecoef\throttle + h_\motorresistancecoef\text{,} \label{eq:throt_to_res}                           \\
    \motorforce      & = a_{\volthrotcoef}\motorvoltage^2 + b_{\volthrotcoef}\motorvoltage\throttle + c_{\volthrotcoef}\throttle^2 + d_{\volthrotcoef}\motorvoltage + e_{\volthrotcoef}\throttle\text{,} \label{eq:volt_throt_to_thrust}
\end{align}
where the quantities $a_{[\cdot]}, b_{[\cdot]}, c_{[\cdot]}, d_{[\cdot]}, e_{[\cdot]}, f_{[\cdot]}, g_{[\cdot]}, h_{[\cdot]}$ are the coefficients of the respective polynomials which can be found in the Table \ref{table:MotorModelCoefficients}, and the plots for the fitted data can be found in Figure \ref{fig:motor_plots}.
\RTwo{The motor-propeller pair is capable of producing up to \SI{20}{\newton} of thrust at \SI{16.8}{\volt} with the bench power supply, but later testing revealed a practical maximum thrust of \SI{17}{\newton} when coupled with the battery, since the battery's internal resistance limits the usable voltage when all four motors are running at full throttle.}

\renewcommand{\arraystretch}{1.3}
\begin{table}[h!]
	\centering
	\begin{tabular}{|c|c|c|}
		\hline
		\textbf{Equation} & \textbf{Coeff.} & \textbf{Value}                   \\
		\hline
        \multirow{3}{*}{$\motorforce(\propvel)$~\eqref{eq:rpm_to_force}}
                          & \( a_{\rpmtoforcecoef} \)         & \( 5.37678924 \times 10^{-8} \)  \\
		                  & \( b_{\rpmtoforcecoef} \)         & \( -1.75562704 \times 10^{-4} \) \\
		                  & \( c_{\rpmtoforcecoef} \)         & \( 2.03427416 \times 10^{-1} \)  \\
		\hline
        \multirow{3}{*}{$\powerin(\motorforce)$~\eqref{eq:thrust_to_power}}
                          & \( a_{\thrusttopowercoef} \)         & \( 1.08195049 \)                 \\
		                  & \( b_{\thrusttopowercoef} \)         & \( 15.24104625 \)                \\
		                  & \( c_{\thrusttopowercoef} \)         & \( -0.5032742180159657 \)        \\
		\hline
        \multirow{8}{*}{$\motorresistance(\throttle)$~\eqref{eq:throt_to_res}}
                          & \( a_{\motorresistancecoef} \)         & \( -1157.75659522 \)             \\
		                  & \( b_{\motorresistancecoef} \)         & \( 5559.67539889 \)              \\
		                  & \( c_{\motorresistancecoef} \)         & \( -11248.33350079 \)            \\
		                  & \( d_{\motorresistancecoef} \)         & \( 12424.55083399 \)             \\
		                  & \( e_{\motorresistancecoef} \)         & \( -8104.5031055 \)              \\
		                  & \( f_{\motorresistancecoef} \)         & \( 3144.94023379 \)              \\
		                  & \( g_{\motorresistancecoef} \)         & \( -688.75561512 \)              \\
		                  & \( h_{\motorresistancecoef} \)         & \( 70.55304472 \)                \\
		\hline
        \multirow{5}{*}{$\motorforce(\motorvoltage, \throttle)$~\eqref{eq:volt_throt_to_thrust}}
                          & \( a_{\volthrotcoef} \)         & \( -1.76279894 \times 10^{-2} \) \\
		                  & \( b_{\volthrotcoef} \)         & \( 1.88742068 \)                 \\
		                  & \( c_{\volthrotcoef} \)         & \( 1.42121264 \times 10^{1} \)   \\
		                  & \( d_{\volthrotcoef} \)         & \( 2.55855147 \times 10^{-1} \)  \\
		                  & \( e_{\volthrotcoef} \)         & \( -2.43087865 \times 10^{1} \)  \\
		\hline
	\end{tabular}
	\caption{Coefficients for fitted polynomials between resistance, thrust, throttle, power, and RPM.}
	\label{table:MotorModelCoefficients}
\end{table}

\eqref{eq:rpm_to_force} represents the aerodynamic property of the propeller and relates the \ac{rpm} of the propeller to the thrust produced by it in static airflow conditions.
\eqref{eq:thrust_to_power} derives from the electrical-to-mechanical power conversion specific to this motor-propeller setup.
According to the model presented by Moseler et al. \cite{moselerApplicationModelbasedFault2000}, there exists a four-way relationship between the voltage, throttle, current, and \ac{rpm}, but it was observed that at higher throttle values, there was very little variation in \ac{rpm} with respect to the voltage.
Therefore, the resulting three-way relationship can be represented by \eqref{eq:throt_to_res}.
Governed by this relationship, the variable resistances shown in Figure \ref{fig:drone_circuit} change as the throttle varies during the flight.
The aforementioned four-way relationship between voltage, throttle, current, and motor \ac{rpm} can also be represented by \eqref{eq:volt_throt_to_thrust}, since the motor \ac{rpm} is directly related to the current of the motor through its torque.

As seen from these plots, various electrical and mechanical properties of this motor-propeller system share a non-linear relationship with one another, and they can be predicted from one another using the aforementioned polynomial functions.
\RThree{All the quantities in these equations represent the physical limits of the actuators except the voltage (which is a property of the battery), and to ensure that they remain valid, these relationships were verified over the entire range of voltages in the flight envelope.}

\begin{figure}
\centering
\includegraphics[width=\linewidth]{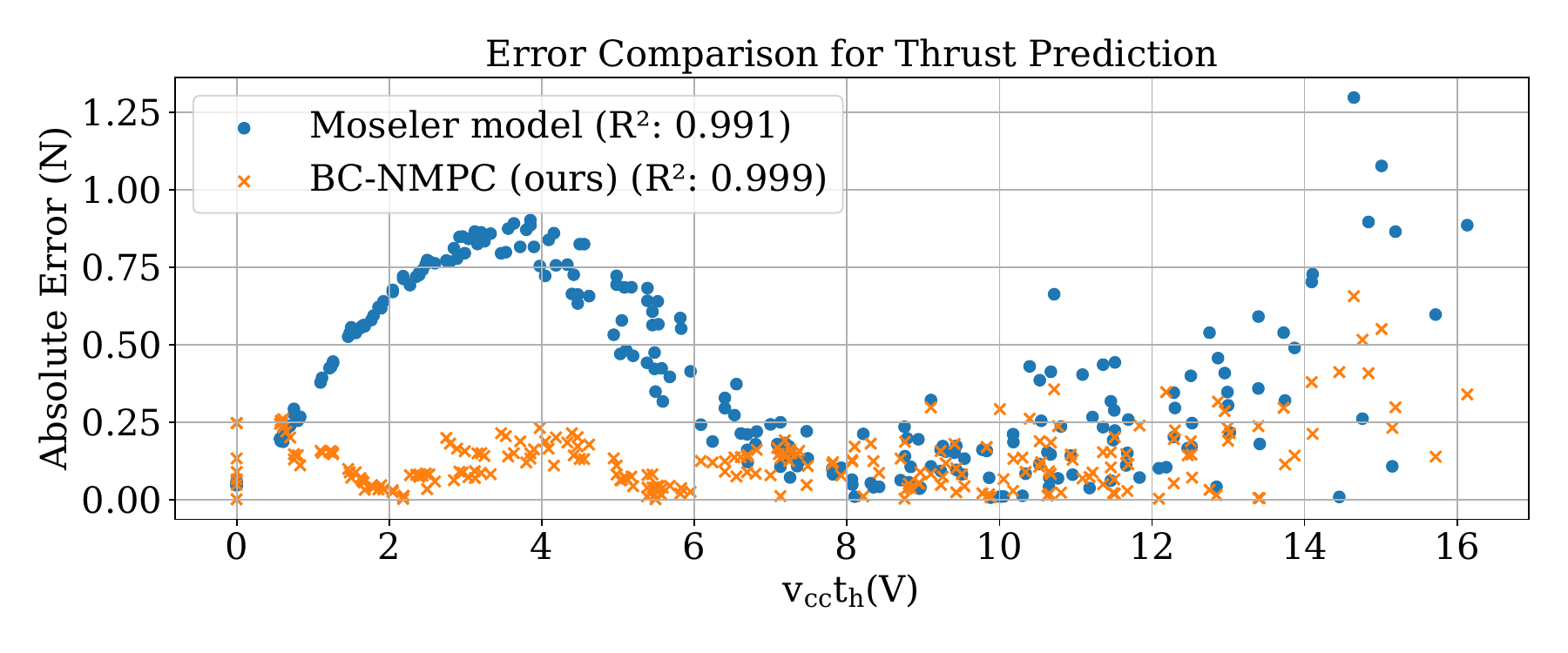}
\caption{Comparison of the thrust predictions from the testing dataset between `BC-NMPC` and the `Moseler model` \cite{moselerApplicationModelbasedFault2000}.}
\label{fig:model_comparison}
\end{figure}

From the fitting dataset obtained through these thrust tests, the parameters for the model proposed by Moseler et al. \cite{moselerApplicationModelbasedFault2000} were also identified, and this model is presented as `Moseler model' in comparison to our proposed model (referred to as `BC-NMPC`).
For any given combination of voltage and throttle, the `Moseler model` can be used to predict the thrust produced by the motor-propeller system, and compared to the thrust predicted by `BC-NMPC` through \eqref{eq:volt_throt_to_thrust}.
\RCommon{The testing dataset from the measurements was used to calculate an absolute error in the predictions of both models, and the results are shown in Figure \ref{fig:model_comparison}.}
\RCommon{As seen from the figure and the coefficient of determination $R^2$, our proposed model performs significantly better than the `Moseler model` for lower throttle ranges and outperforms it for higher throttle ranges as well.}

\begin{figure}
    \centering
    \includegraphics[width=0.35\textwidth, angle=90]{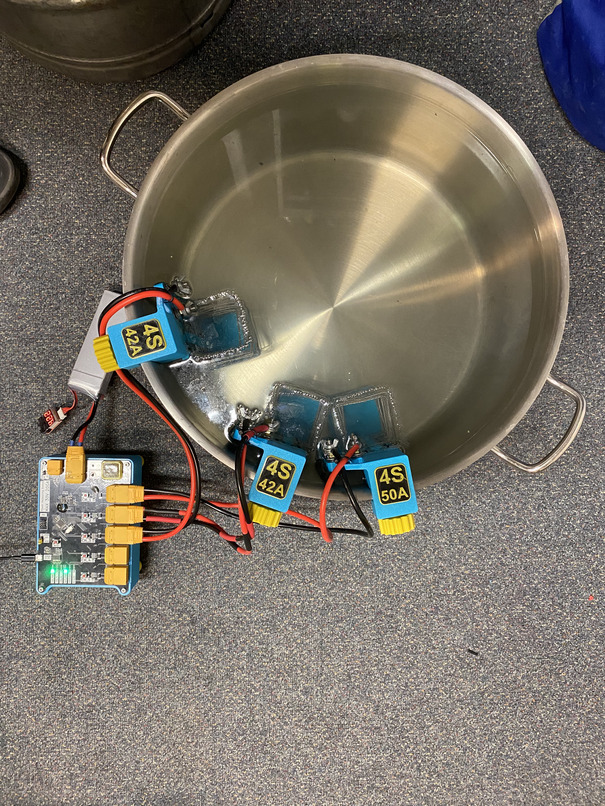}
    \caption{Battery testing setup used to obtain the battery discharge characteristics.}
    \label{fig:battery_tester}
\end{figure}

\subsection{Battery model}
\label{sec:battery_model}
The equivalent \ac{ir} circuit model consists of an ideal voltage source representing open circuit voltage $\batteryvoltage$ and a resistor $\batteryresistance$ representing the battery's internal resistance, as seen in Figure \ref{fig:drone_circuit}.
The parameters of the \ac{ir} model have been shown to be dependent on the \ac{soc} of the battery $\batterysoc$ \cite{Battery_modelling, Doerffel2006}, and therefore, discharge tests were conducted to obtain the relations between $\batteryvoltage$, $\batteryresistance$, and $\batterysoc$.

These tests were conducted using a custom-designed battery discharge setup (see Figure \ref{fig:battery_tester}) which could discharge the battery in multiples of \SI{40}{\ampere}. 
For our agile \ac{uav}, and therefore, for this test, we used a Gaoneng GNB 4S \SI{3.3}{\ampere\hour} \ac{lipo} battery rated at 100C discharge with an XT90 connector.
Based on the flight telemetry for our aircraft during \SI{3.5}{\accelgrav} trajectories, a discharge current of \SI{120}{\ampere} was chosen to reflect the maximum discharge load in flight.
This testing setup may not always be necessary as the manufacturer's datasheets often provide sufficient technical data for parameter identification.

\begin{figure}
	\centering
	\begin{subfigure}[b]{0.5\textwidth}
		\centering
		\includegraphics[width=\linewidth]{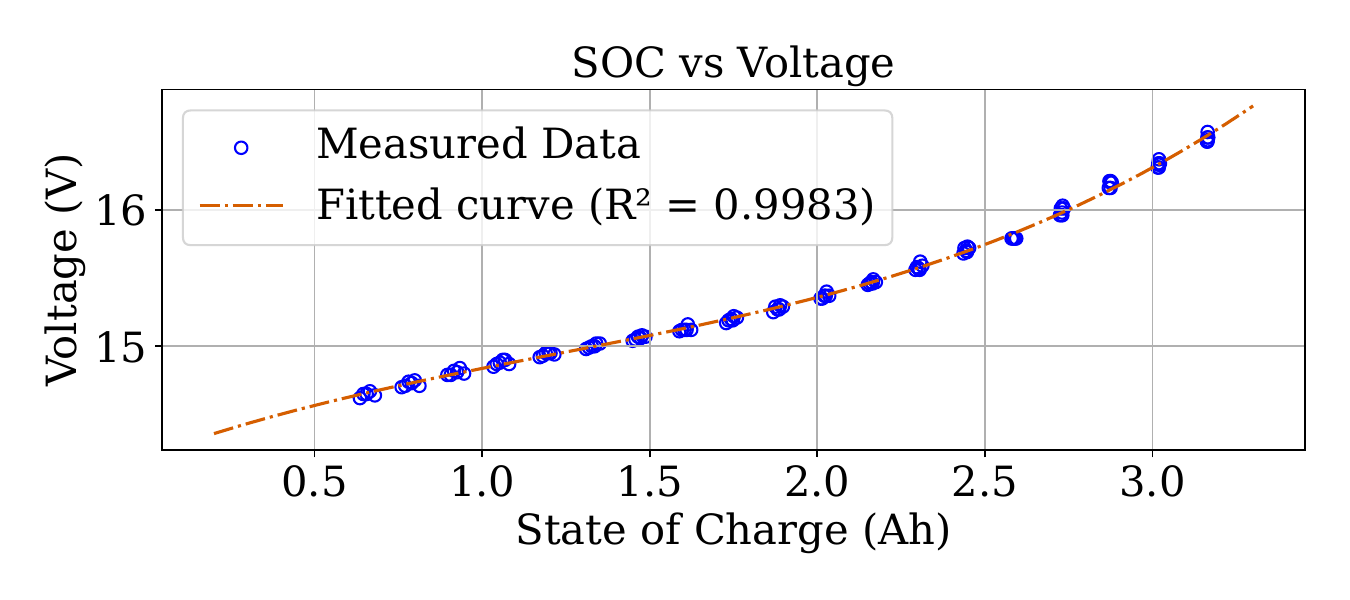}
		\label{fig:soc_vs_volt}
	\end{subfigure}
	\begin{subfigure}[b]{0.5\textwidth}
		\centering
		\includegraphics[width=\linewidth]{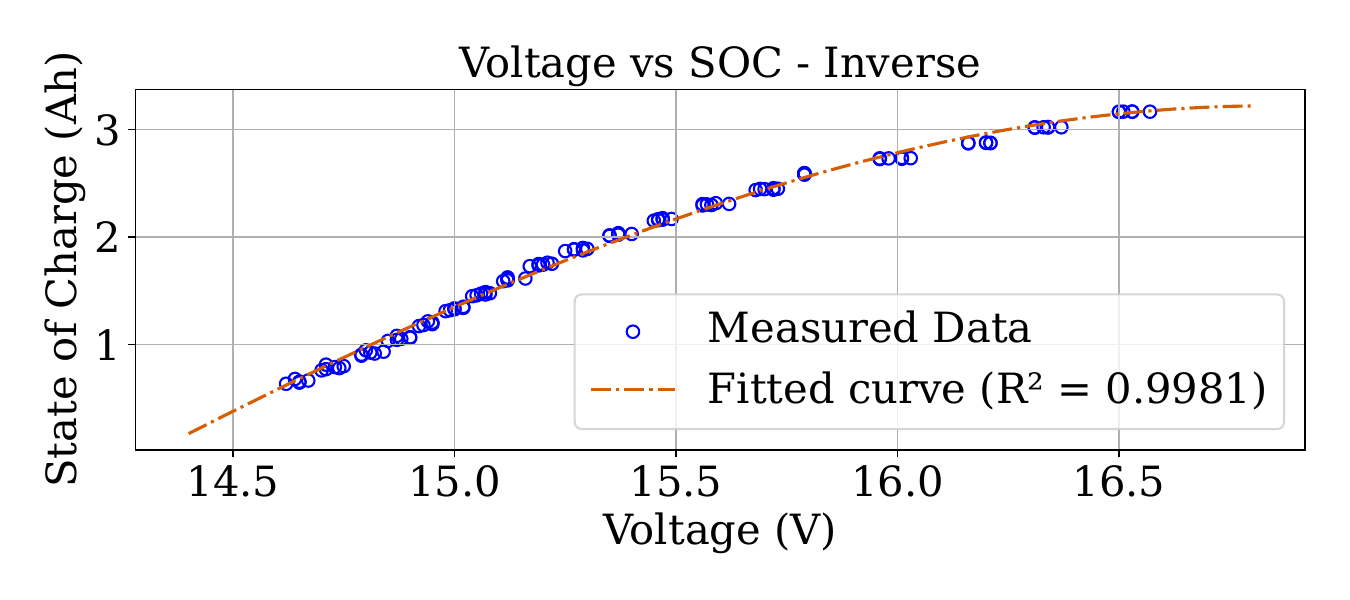}
		\label{fig:volt_vs_soc}
	\end{subfigure}
	\begin{subfigure}[b]{0.5\textwidth}
		\centering
		\includegraphics[width=\linewidth]{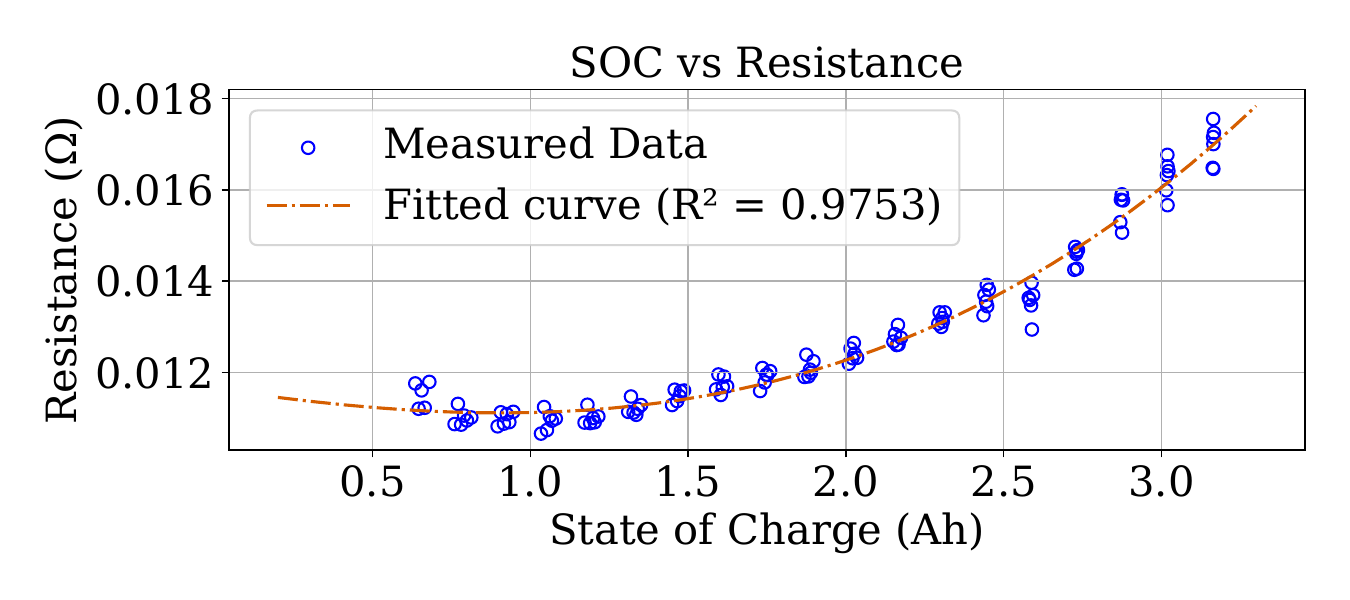}
		\label{fig:soc_vs_resistance}
	\end{subfigure}

	\caption{Relations between open circuit voltage, internal resistance and state of charge.}
	\label{fig:battery_plots}
\end{figure}

Using the polynomial fitting of the data obtained by measuring battery discharge, relationships between open circuit voltage $\batteryvoltage$, state of charge $\batterysoc$, and internal resistance $\batteryresistance$ were obtained such that

\begin{align}
    \batterysoc &= a_{\volttosoccoef}\batteryvoltage^3 + b_{\volttosoccoef}\batteryvoltage^2 + c_{\volttosoccoef}\batteryvoltage + d_{\volttosoccoef}\text{,} \label{eq:bat_soc}\\
    \batteryresistance &= a_{\soctorescoef}\batterysoc^3 + b_{\soctorescoef}\batterysoc^2 + c_{\soctorescoef}\batterysoc + d_{\soctorescoef}\text{, and} \label{eq:bat_res}\\
    \batteryvoltage &= a_{\soctovoltcoef}\batterysoc^3 + b_{\soctovoltcoef}\batterysoc^2 + c_{\soctovoltcoef}\batterysoc + d_{\soctovoltcoef}, \label{eq:bat_volt}
\end{align}

where the quantities $a_{[\cdot]}, b_{[\cdot]}, c_{[\cdot]}$ are the coefficients of the respective polynomials which can be found in the Table \ref{table:CombinedBatteryCoefficients}, and the plots for the fitted data can be found in Figure \ref{fig:battery_plots}.
The obtained polynomials can be used to efficiently predict the battery states in real-time, and are further integrated into a \ac{nmpc} formulation as shown in the next section.
\renewcommand{\arraystretch}{1.3}
\begin{table}[h!]
	\centering
	\begin{tabular}{|c|c|c|}
		\hline
		\textbf{Equation} & \textbf{Coeff.} & \textbf{Value}                   \\
		\hline
        \multirow{4}{*}{$\batterysoc(\batteryvoltage)$~\eqref{eq:bat_soc}}
                          & \( a_{\volttosoccoef} \)         & \( -6.95090382 \times 10^{-2}\)                \\
		                  & \( b_{\volttosoccoef} \)         & \( 2.82697462 \)                 \\
		                  & \( c_{\volttosoccoef} \)         & \( -36.0852050 \)                \\
		                  & \( d_{\volttosoccoef} \)         & \( 141.150339 \)                 \\
		\hline
        \multirow{4}{*}{$\batteryresistance(\batterysoc)$~\eqref{eq:bat_res}}
                          & \( a_{\soctorescoef} \)         & \( 1.60148451 \times 10^{-4} \) \\
		                  & \( b_{\soctorescoef} \)         & \( 3.55487546 \times 10^{-4} \) \\
		                  & \( c_{\soctorescoef} \)         & \( -1.03989599 \times 10^{-3} \) \\
		                  & \( d_{\soctorescoef} \)         & \( 0.0116445337 \)               \\
		\hline
        \multirow{4}{*}{$\batteryvoltage(\batterysoc)$~\eqref{eq:bat_volt}}
                          & \( a_{\soctovoltcoef} \)         & \( 9.22112995 \times 10^{-2} \) \\
		                  & \( b_{\soctovoltcoef} \)         & \( -3.37933271 \times 10^{-1} \) \\
		                  & \( c_{\soctovoltcoef} \)         & \( 8.88619439 \times 10^{-1} \)  \\
		                  & \( d_{\soctovoltcoef} \)         & \( 14.1950250 \)                 \\
		\hline
	\end{tabular}
	\caption{Polynomial coefficients for relations between $\batterysoc$, $\batteryvoltage$ and $\batteryresistance$.}
	\label{table:CombinedBatteryCoefficients}
\end{table}

\section{Battery-Constrained \acf{nmpc}}
\label{sec:bc_nmpc_description}

This section presents the employed methodology to integrate the propulsion model into the standard \ac{nmpc} formulation for real-time prediction and control.
Following subsections discuss the control problem formulation and its associated constraints for the `BC-NMPC` model.

\subsection{\ac{uav} system dynamics model}\label{sec:uav_model}
Similar to our previous work \cite{11246583} and other works in the literature, a \ac{uav}'s system dynamics model can be described using position $\positionvector \in \Rthree$, velocity $\linearvelvector \in \Rthree$, \RThree{unit quaternion $\orientation \in S^3$}, body rates $\angularvelvector \in \Rthree$, and battery's \ac{soc} $\batterysoc\in \mathbb{R}$.

To formulate this system into an \ac{ocp}, state vector of the \ac{uav} can be defined as $\uavstate = [\positionvector~\linearvelvector~\orientation~\angularvelvector~\batterysoc]^\top$, and the input vector $\uavinput$ can be chosen such that $\uavinput = \thrustvector: [\motorforce_1~\motorforce_2~\motorforce_3~\motorforce_4]^\top$, where $\motorforce_i$ is the scalar thrust force produced by each of the four motors, and $\thrustvector$ is the vector of thrusts produced.
The derivatives of the \ac{uav} states are then given as
\begin{align}
    \small
	\label{eq:quad_dyn}
	\begin{aligned}
		\bm{\dot{\positionvector}} & = {\linearvelvector} \text{,} \vphantom{\frac{1}{2}}                           \\
		\bm{\dot{\linearvelvector}} & = \frac{\bm{R}(\orientation)(\bodythrustvector+\dragforcevector)}{m} + \bm{g}\text{,} \\
        \bm{\dot{\orientation}}      & = \frac{1}{2} \orientation \circledast \begin{bmatrix} 0 \\ \angularvelvector \end{bmatrix}\text{,}     \\
        \bm{\dot{\angularvelvector}} & = \inertiamat^{-1} ({\torquevector} - \bm{\omega} \times \inertiamat \bm{\omega})\text{,} \\
        \dot{s}_{oc} & = -\circuitcurrent/3600\text{,} 
	\end{aligned}
\end{align}
where the operator $\circledast$ denotes the quaternion multiplication, $\bm{R}(\orientation)$ is the rotational matrix of quaternion $\orientation$, $\bodythrustvector$ is the thrust vector in the body frame, $\dragforcevector$ is the drag force vector in the body frame, $m$ is the mass of the UAV, $\bm{g}$ is the gravitational acceleration, $\inertiamat$ is the diagonal inertial matrix of the rigid body of the \ac{uav}, $\torquevector$ is the torque produced in the body frame, and $\circuitcurrent$ is the current drawn from the battery.

Drag force $\dragforcevector$ in~\eqref{eq:quad_dyn} is usually modelled as a linear function of velocity $\linearvelvector_{B}$ in body frame ${B}$  with drag coefficients $(k_{vx},k_{vy},k_{vz})$ \cite{Sun2022NMPCcomparativeStudy}, and can be computed as
\begin{equation}
	\normalsize
	\label{eq:drag}
	\dragforcevector = - \begin{bmatrix}k_{vx} v_{B,x} & k_{vy} v_{B,y} & k_{vz} v_{B,z}\end{bmatrix}^{\top} \text{.}
\end{equation}

The body thrust vector $\bodythrustvector = [0~0~\collectivethrust]^\top,$ and the torque vector $\torquevector$ are computed using allocation matrix such that
\begin{equation}
	\normalsize
	\begin{aligned}
		\label{eq:motor_allocation}
		\begin{bmatrix} \collectivethrust \\ \torquevector \end{bmatrix}
		                  & =
		\begin{bmatrix}
			1           & 1            & 1            & 1          \\
			l/\sqrt{2}  & - l/\sqrt{2} & - l/\sqrt{2} & l/\sqrt{2} \\
			-l/\sqrt{2} & -l/\sqrt{2}  & l/\sqrt{2}   & l/\sqrt{2} \\
			\kappa      & -\kappa      & \kappa       & -\kappa
		\end{bmatrix}
		\thrustvector\text{,}
	\end{aligned}
\end{equation}
where $l$ is the arm length of the symmetric \ac{uav} frame, and $\kappa$ is the motor torque constant.

At any given timestep, the $\dot{s}_{oc}$ is computed as a function of the control input $\uavinput$ by describing the relationship between desired thrust and the induced current in the system.
From \eqref{eq:thrust_to_power}, the total power consumed by all motors ($\powerin$) is given by the sum of power consumed by each motor such that
\begin{equation}
	\label{total_power}
	\powerin = \sum^4_{i=1} \powerinsingle =\sum^4_{i=1} \left(a_{\thrusttopowercoef}\motorforcesingle^2 + b_{\thrusttopowercoef}\motorforcesingle + c_{\thrusttopowercoef}\right)\text{.}
\end{equation}
Using \eqref{eq:volt_div} , the voltage and the power can be expressed as 
\begin{equation}
	\label{eq:current_eq}
	\circuitcurrent^2\batteryresistance - \circuitcurrent \batteryvoltage + \powerin = 0\text{,}
\end{equation}
with discriminant
\begin{equation}
	\label{eq:discriminant}
	D = \batteryvoltage^2 - 4\powerin\batteryresistance\text{.}
\end{equation}
{
This quadratic equation may have one, two, or no real solutions, depending on the value of the discriminant. 
The discriminant of the equation represents the maximum permissible power draw ($p_{max} = \batteryvoltage^2/4\batteryresistance$) from the battery during its entire discharge cycle, and this power draw varies as the battery discharges.
For the solution to be real and feasible, $\powerin \leq p_{max}$ must hold, and so, using \eqref{eq:bat_res} and \eqref{eq:bat_volt}, the lowest possible value of $p_{max}$ was found to be $\approx \SI{4000}{\watt}$.
From \eqref{eq:thrust_to_power} and Figure \ref{fig:motor_plots}, it can be seen that an individual motor can only draw a maximum of $\approx \SI{800}{\watt}$, and therefore, the maximum power consumption of the system is $\approx \SI{3200}{\watt}$, which is lower than the lowest possible value of $p_{max}$.
Hence, two roots of the equation exist in the entire flight envelope, and from these two roots of the equation, only the negative root was found to be a valid root for the current ranges possible in the system.
Using \eqref{eq:current_eq}, the current drawn from the battery can be computed in flight to predict the change in \ac{soc} of the battery.
}

This presented dynamics model is used to predict the state of the \ac{uav} at the next timestep, given the current state and the control input, in the \ac{nmpc} formulation as shown in the next section.

\subsection{BC-NMPC formulation}\label{sec:ocp_formulation}
For this work, the \ac{nmpc} is implemented as an \ac{ocp} using \textit{ACADOS} \cite{Acados} framework.
For this \ac{ocp}, the state vector, the input vector, and the system model are defined as stated in Section \ref{sec:uav_model}.
The formulation follows the standard \ac{nmpc} structure used in previous work \cite{Sun2022NMPCcomparativeStudy, 11246583}:
{
\begin{equation}
    \normalsize
	\begin{aligned}
		\argmin_{\uavinput_0,\ldots,\uavinput_{\mpchorizon}} J(&\statestd, \uavinput)  = \sum_{k = 1}^{\mpchorizon}\errormat_k^T \errpenmat \errormat_k + \sum_{k = 0}^{\mpchorizon-1}\errorinputmat_{k}^T \inputpenmat \errorinputmat_{k}\\ \text{subject to :} \\
        \errormat_k                                                               & = \statestddesired_k - \statestd_k,~ \errorinputmat_k = \uavinputdesired_k - \uavinput_k,~  \\
		\statestd_0                                                               & = \statestd_\text{init},~ 
        \statestd_{k+1}                                                            = f_\text{RK4}(\statestd_k,\uavinput_k), \\
        \uavinput_k                                                               & \in [\thrustvector_\text{min},\thrustvector_\text{max}],~\text{(control input const.)} \\
        \angularvelvector_k                                                        & \in [\angularvelvector_\text{min},\angularvelvector_\text{max}]\text{, (state const.)}    \\
        \underline{\bm{h}} &\leq \bm{h}(\uavstate_k, \uavinput_k) \leq \overline{\bm{h}},~\text{(nonlin. const.)} \label{eq:nonlinear_constraints} 
	\end{aligned}
\end{equation}
where $[\bullet]_k$ is the quantity at $t = t_\text{init} + k\Delta t$, $\errormat_k$ is the error in the state, $\statestddesired_k$ is the desired (reference) state, $\errorinputmat_k$ is the deviation of the input from the reference $\uavinputdesired_k$, $\mpchorizon$ is the prediction horizon, $\errpenmat$ and $\inputpenmat$ are the positive definite matrices for the state and input penalties respectively, $\thrustvector_\text{min}$ and $\thrustvector_\text{max}$ are the control input limits for individual actuators independently, and finally, $\angularvelvector_\text{min}$ and $\angularvelvector_\text{max}$ represent the limits on body rates of the aircraft.
$\errpenmat$ is a diagonal matrix comprised as diag($\errpenmat_{\positionvector}, \errpenmat_{\quaternion}, \errpenmat_{\linearvelvector}, \errpenmat_{\angularvelvector}$), and the values for these penalties as well as the values for other parameters of the model are given by Table \ref{table:OtherCoefficients}.
}

\begin{table}[h!]
	\centering
	\begin{tabular}{|l|l|} 
		\hline
		\textbf{Coeff.} & \textbf{Value} \\
		\hline
        $\errpenmat_{\positionvector}$ & $100$ \\
        $\errpenmat_{\quaternion}$ & $26$ \\
        $\errpenmat_{\linearvelvector}$ & $6 $ \\
        $\errpenmat_{\angularvelvector}$ & $26$ \\
        $\inputpenmat$ & diag($4,4,4,4$) \\
        $\mpchorizon$ & $20$ \\
        $\Delta t~(\unit{\milli\second})$ & $50$ \\
		\hline
        $\angularvelvector_{\text{min}}~(\unit{\radian\per\second})$ & $[-6~-6~-3]^T$ \\
        $\angularvelvector_{\text{max}}~(\unit{\radian\per\second})$ & $[6~6~3]^T$ \\
        $\thrustvector_{\text{min}}$~(\unit{\newton}) & $[0.5~0.5~0.5~0.5]^T$ \\
        $\thrustvector_{\text{max}}$~(\unit{\newton}) & $[17~17~17~17]^T$ \\
        $\inertiamat_{xx}~(\unit{\kilo\gram\square\meter})$ & $0.007$ \\
        $\inertiamat_{yy}~(\unit{\kilo\gram\square\meter})$ & $0.007$ \\
        $\inertiamat_{zz}~(\unit{\kilo\gram\square\meter})$ & $0.011$ \\
        $l~(\unit{\meter})$ & $0.015$ \\
        $\kappa~(\unit{\meter})$ & $0.012$ \\
        $k_{vx}~(\unit{\newton\second\per\metre})$ & $0.28$ \\
        $k_{vy}~(\unit{\newton\second\per\metre})$ & $0.35$ \\
        $k_{vz}~(\unit{\newton\second\per\metre})$ & $0.7$ \\
		\hline
	\end{tabular}
    \caption{Model Coefficients}
    \label{table:OtherCoefficients}
\end{table}

As a consequence of the internal resistance of the battery being in series with the parallel network of the motors, the current consumed by each actuator affects the available power to other actuators through the induced voltage drop. 
When all four motors are operating at full throttle, the induced voltage drop reduces the maximum power each motor can draw, leading to a reduction in the maximum thrust that each motor can produce.
The maximum thrust available for the system also changes over time as the internal resistance and open circuit voltage of the battery vary with the \ac{soc} of the battery.

The maximum available collective thrust $\collectivethrust_\text{max}$ (at $\throttle=1$) can be calculated using \eqref{eq:volt_throt_to_thrust} and the voltage $\motorvoltage$ across the motor network, which is a function of the battery voltage $\batteryvoltage$, battery resistance $\batteryresistance$, and the total power consumed by the motors $\powerin$.
\RCommon{Circuit voltage $\circuitvoltage$ can be calculated from \eqref{eq:volt_div} after finding the the individual motor resistance using \eqref{eq:throt_to_res}.}
$\collectivethrust_\text{max}$ at this $\motorvoltage$ can be calculated using \eqref{eq:volt_throt_to_thrust} such that
\begin{equation}
	\label{eq:T_max}
    \collectivethrust_{\text{max}} =  4 \left(a_{\volthrotcoef}\motorvoltage^2 + b_{\volthrotcoef}\motorvoltage + d_{\volthrotcoef}\motorvoltage + c_{\volthrotcoef} + e_{\volthrotcoef}\right) \text{.}
\end{equation}

Since \eqref{eq:T_max} is non-linear with respect to the states of the \ac{uav}, the non-linear constraint $\bm{h}(\uavstate_k, \uavinput_k)$ in \eqref{eq:nonlinear_constraints} is defined at each timestep as:
\begin{equation}
	\label{eq:h_constr}
	0 \leq \collectivethrust_{\text{max,k}} - \sum_{i = 1}^4 \motorforce_{i,k}\text{,}
\end{equation}
where $\collectivethrust_{\text{max,k}}$ is the collective thrust limit given by the \eqref{eq:T_max} at $k^\text{th}$ timestep, and $\motorforce_{i, k}$ is the thrust generated by the $i^\text{th}$ motor at the $k^\text{th}$ timestep. 
This constraint ensures that the desired collective thrust input at each time step does not exceed the capabilities of the \ac{uav} given the current \ac{soc} of the battery.

\RCommon{The independent rotor thrust of a single actuator is set conservatively to \SI{17.0}{\newton} during flight to account for the loss of thrust due to battery resistance when all actuators are set to full throttle. 
In this work, we do not formulate a variable constraint on $\thrustvector_\text{max}$ as a function of the $\batterysoc$ of the battery due to the cascaded nature of the control stack wherein the desired body rates are sent to the low-level controller, and the torque allocation is performed by the low-level stack.
Upon investigation, it was found that the desired torque inputs were always met by the low-level controller for the aforementioned limit per actuator and so, the effect of independent single-rotor thrust depletion is minimised through this choice.}

\section{Bench Test Confirmation}\label{sec:max_bench_verification}
\begin{figure}
    \centering
    \begin{subfigure}[b]{\linewidth}
        \centering
        \includegraphics[width=\linewidth]{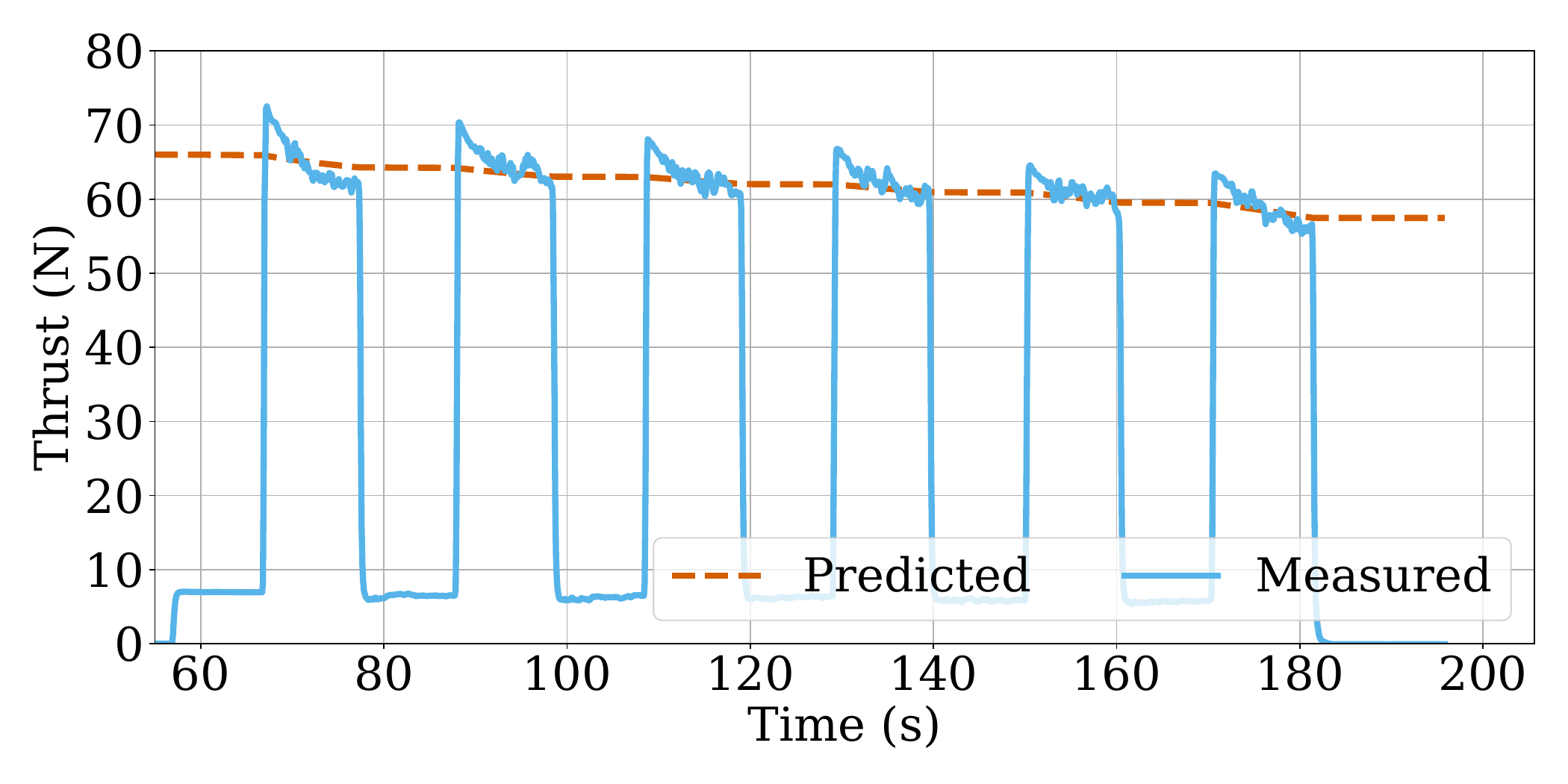}
        \caption{Maximum thrust predictions for the \ac{uav} at $100\%$ throttle compared to the measured thrust.}
        \label{fig:max_thrust_validation}
    \end{subfigure}
    \begin{subfigure}[b]{\linewidth}
        \centering
        \includegraphics[width=\linewidth]{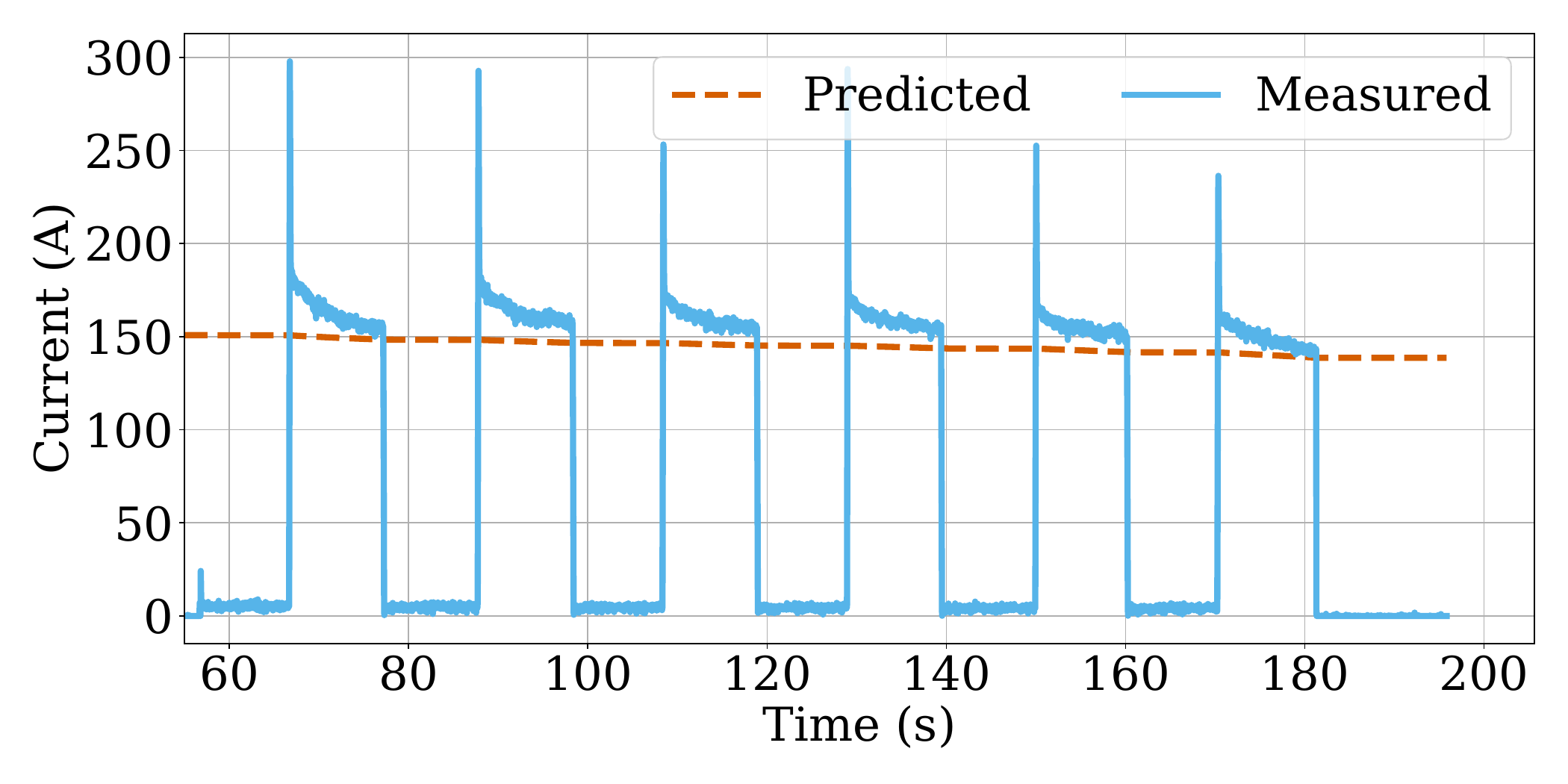}
        \caption{Current predictions at maximum thrust for the \ac{uav} at $100\%$ throttle compared to the measured current.}
        \label{fig:max_current_validation}
    \end{subfigure}
    \caption{Validation of maximum thrust and current predictions at maximum thrust using a static thrust test.}
    \label{fig:max_thrust_current_validation}
\end{figure}

For the purpose of the presented work, it is crucial to validate the model's ability to correctly predict the maximum available thrust of the \ac{uav} at any given time.
While the initial static thrust tests were performed using only a single motor-propeller unit with bench power supply, the full model of propulsion was verified using a thrust stand with the entire \ac{uav} powered by the battery.
For this experiment, the \ac{uav} was cycled between $10\%$ and $100\%$ throttle, and the resulting thrust and current were measured using the thrust stand and the onboard power module, respectively.
Using the methodology presented in Section \ref{sec:ocp_formulation}, the maximum available thrust and the circuit current were predicted for maximum throttle and were compared against the telemetry as shown in Figure \ref{fig:max_thrust_current_validation}.
As seen in the plots, the predictions for both the maximum thrust and the circuit current align well with the measurements at $100\%$ throttle.
One thing to note are the large spikes in current which occurred both with bench power supply and the battery.
These spikes are a common property of the trapezoidal pulse control of \ac{bldc} motors and they are caused by the sudden change of supply voltage in motor control.

\RCommon{While this validation alone is sufficient to lay the foundation for trajectory replanning based on predicted $\collectivethrust_\text{max}$, we performed further rigorous testing of the model's ability to predict the propulsion states for highly agile and dynamic flight scenarios for drone racing applications.}

\section{Real World Experiments}
The \ac{ocp} formulation containing our proposed model and its underlying non-linear constraint (defined in Section \ref{sec:ocp_formulation}) is required to be solved onboard a \ac{uav} at \SI{100}{\hertz}, and therefore, we performed real-world experiments to validate the computational feasibility of the \ac{nmpc} formulation with the integrated real-time thrust constraint and the propulsion predictions of our proposed model.
For this purpose, we tested our model in a very challenging and dynamic scenario by flying a custom-built \ac{uav} in an agile high-speed trajectory at $\SI{3.5}{\accelgrav}$ of acceleration.

The quadrotor measures \SI{300}{\milli\meter} diagonally (motor-to-motor) and weighs \SI{1.2}{\kilo\gram}, as shown in Figure \ref{fig:intro_pic}. 
\RCommon{Due to the uncertain nature of these experiments where the \ac{uav} was flown to the end of the battery limit, the mass was lowered from the usual \SI{1.54}{\kilo\gram} by removing the onboard Livox LiDAR, in order to prevent loss of valuable instruments in the event of a crash.}
For low-level control, the \ac{uav} is equipped with a CubePilot Cube Orange+ flight controller running \texttt{PX4} firmware and is commanded through the open-source MRS system architecture \cite{Baca2021} running on a Khadas Vim3 Pro single-board computer. 
For state estimation, the \ac{uav} is equipped with a Holybro F9P RTK GNSS module and receives real-time corrections from a base station, which are fused with the \ac{imu} onboard the flight controller to produce odometry for the system.
The electric and mechanical propulsion components of the \ac{uav} are the same as described in the test setups in Sections \ref{sec:motor_model} and \ref{sec:battery_model}.

\RCommon{For the experiments, the onboard computer solved the aforementioned \ac{ocp} every \SI{10}{\milli\second}, and the first predicted body rate vector $\angularvelvector_1$ was sent to the underlying \texttt{PX4} stack.}
Crucially, the desired throttle for the aircraft was calculated from an inverse of \eqref{eq:volt_throt_to_thrust} using the desired collective thrust input and the instantaneous terminal voltage $\motorvoltage$.
\RCommon{At boot-up, the open-circuit voltage $\batteryvoltage$ was used to find the initial \ac{soc} using the polynomial \eqref{eq:bat_soc}.}
During the flight, $\motorvoltage$ and $\circuitcurrent$ were measured using the onboard power module at a frequency of \SI{100}{\hertz}, and the \ac{soc} was updated using the Coulomb counting method.
\begin{figure}
    \centering
    \includegraphics[width=0.5\textwidth]{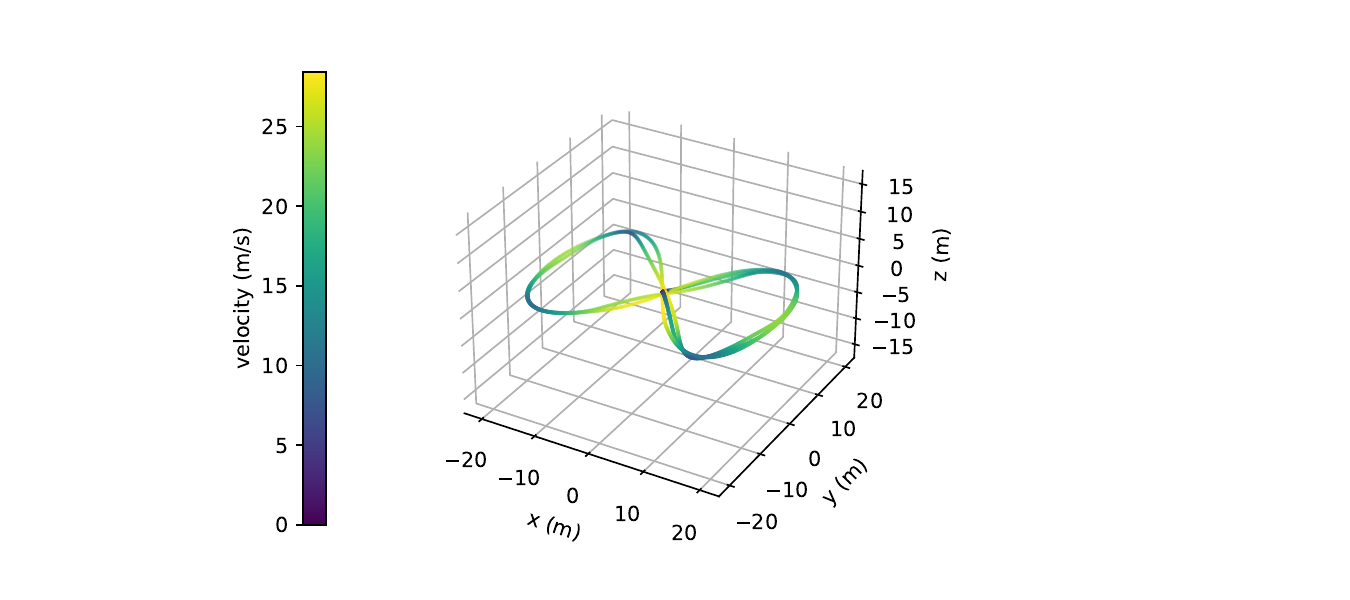}
    \caption{Trajectory flown during real-world experiments.}
    \label{fig:fig8_trajectory}
\end{figure}
The \ac{uav} flew a pre-defined series of waypoints in a lemniscate trajectory (see Figure \ref{fig:fig8_trajectory}) generated using work presented in \cite{PMM}. 
\RCommon{This trajectory was chosen for model validation since it features a combination of sharp turns at high-speeds.}
The \ac{uav} began flight with a nearly fully-charged \ac{lipo} battery at \SI{3.3}{\ampere\hour}, and flew the same waypoints repeatedly until the battery was nearly depleted at an \ac{soc} of approximately \SI{0.3}{\ampere\hour} at the end.

We highlight that in comparison to the literature discussed in Section \ref{sec:system_model}, and especially in comparison to work presented in \cite{bauersfeld2022range}, our experimental setup features a highly dynamic trajectory with highly varying speeds and accelerations that are representative of the real-world applications of agile \acp{uav}.
The aircraft achieved a maximum velocity of \SI{23}{\meter\per\second} and a maximum acceleration of \SI{44}{\meter\per\second\squared} during the flight.

In the following sections, we discuss the comparison between our estimated, measured, and ground truth values.
For the discussion of results, we use $ \estcircuitvoltage, \estcircuitcurrent, \estbatterysoc, \estbatteryresistance, \estbatteryvoltage$ to denote estimated quantities, and $ \meascircuitvoltage, \meascircuitcurrent, \measbatteryresistance$ to denote measured quantities. 
Using the methodology presented in Sections \ref{sec:motor_model} , \ref{sec:battery_model} , \ref{sec:ocp_formulation}, $\estcircuitcurrent, \estbatterysoc, \estbatteryresistance, \estbatteryvoltage$ can be estimated.
$\estcircuitvoltage$ can then be estimated indirectly as discussed in the relevant section below.

Among the measured quantities, $\meascircuitvoltage$ is a low-noise highly accurate measurement while $\meascircuitcurrent$ is a high noise measurement.
On the other hand, $\measbatteryresistance$ is an indirect measurement made from other measurements, $\batterysoc$ cannot be measured, and finally, $\batteryvoltage$ can only be measured at zero current draw which usually occurs at the beginning or the end of a flight.
\subsection{Open-circuit voltage validation}
For the validation of open-circuit voltage estimate $\estbatteryvoltage$, the initial open-circuit voltage of the battery was measured when the current drawn from the battery was zero, since for $\circuitcurrent = 0$, $\circuitvoltage = \batteryvoltage$.
This measurement was used to initialise the model with \ac{soc} of the battery using the \eqref{eq:bat_soc}.
During the flight, the prediction model was constantly updated using Coulomb counting and the final $\batteryvoltage (=\circuitvoltage)$ was then re-measured post-landing to confirm the accuracy of the $\estbatteryvoltage$ estimation.

\begin{figure}
\centering
\includegraphics[width=\linewidth]{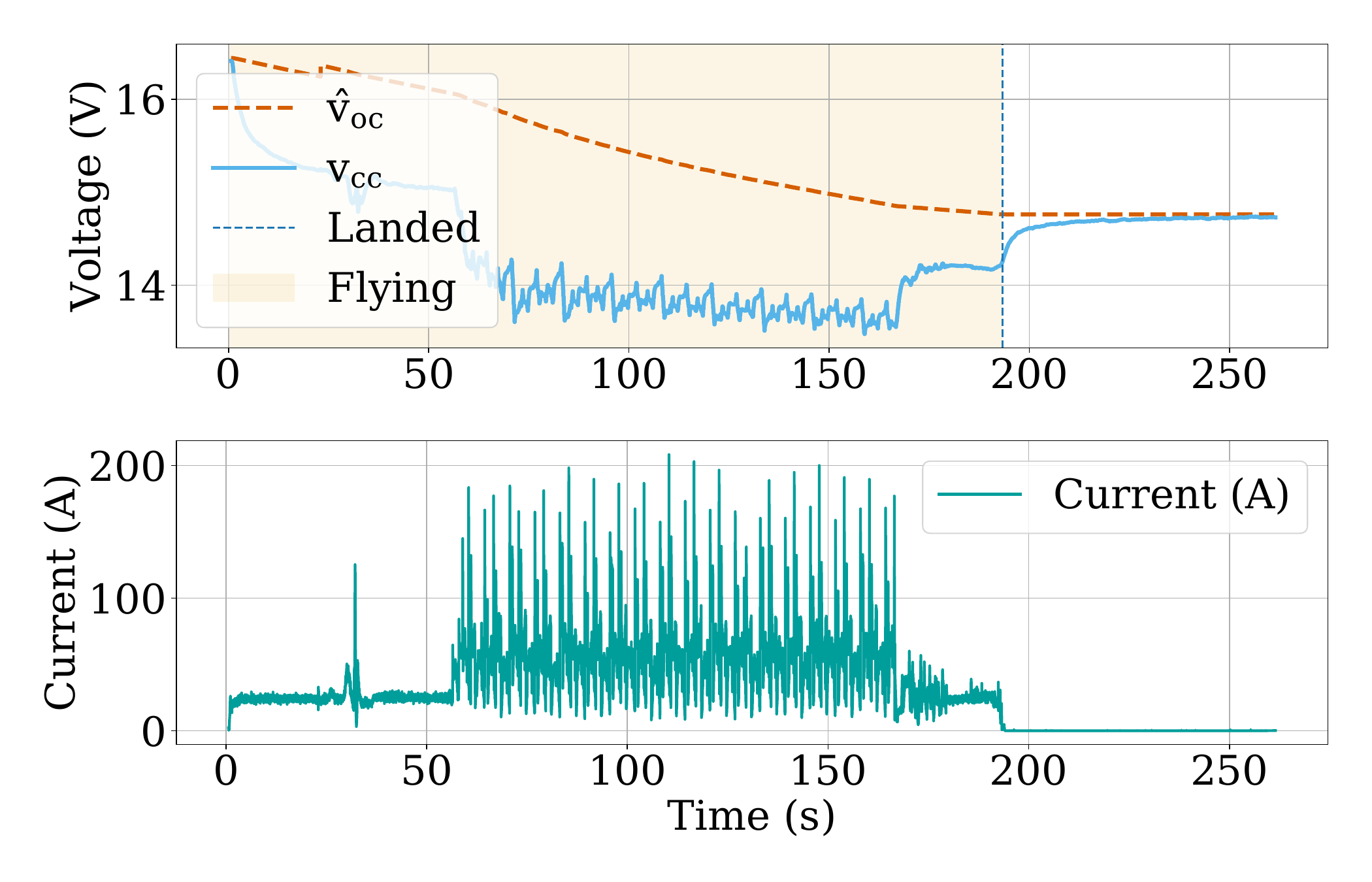}
\caption{Comparison of measured terminal voltage $\circuitvoltage$ and estimated open-circuit voltage $\estbatteryvoltage$.}
\label{fig:oc_voltage}
\end{figure}

A comparison between the resulting $\circuitvoltage$ and the estimated open-circuit voltage $\estbatteryvoltage$ is shown in Figure~\ref{fig:oc_voltage}.

As shown in the plot, $\circuitvoltage$ and $\estbatteryvoltage$ converge during the idle period post-landing, indicating that the open-circuit voltage can be reliably predicted even after an aggressive flight.
It also confirms the Coulomb counting approach and its ability to estimate the $\estbatterysoc$ throughout the flight.
These reliable estimates of $\estbatteryvoltage$ and $\estbatterysoc$ form a solid foundation for the validation of the internal resistance estimation, as shown in the following section.
\subsection{Internal Resistance Estimation}
\begin{figure}
    \centering
    \begin{subfigure}[b]{\linewidth}
        \centering
        \includegraphics[width=\linewidth]{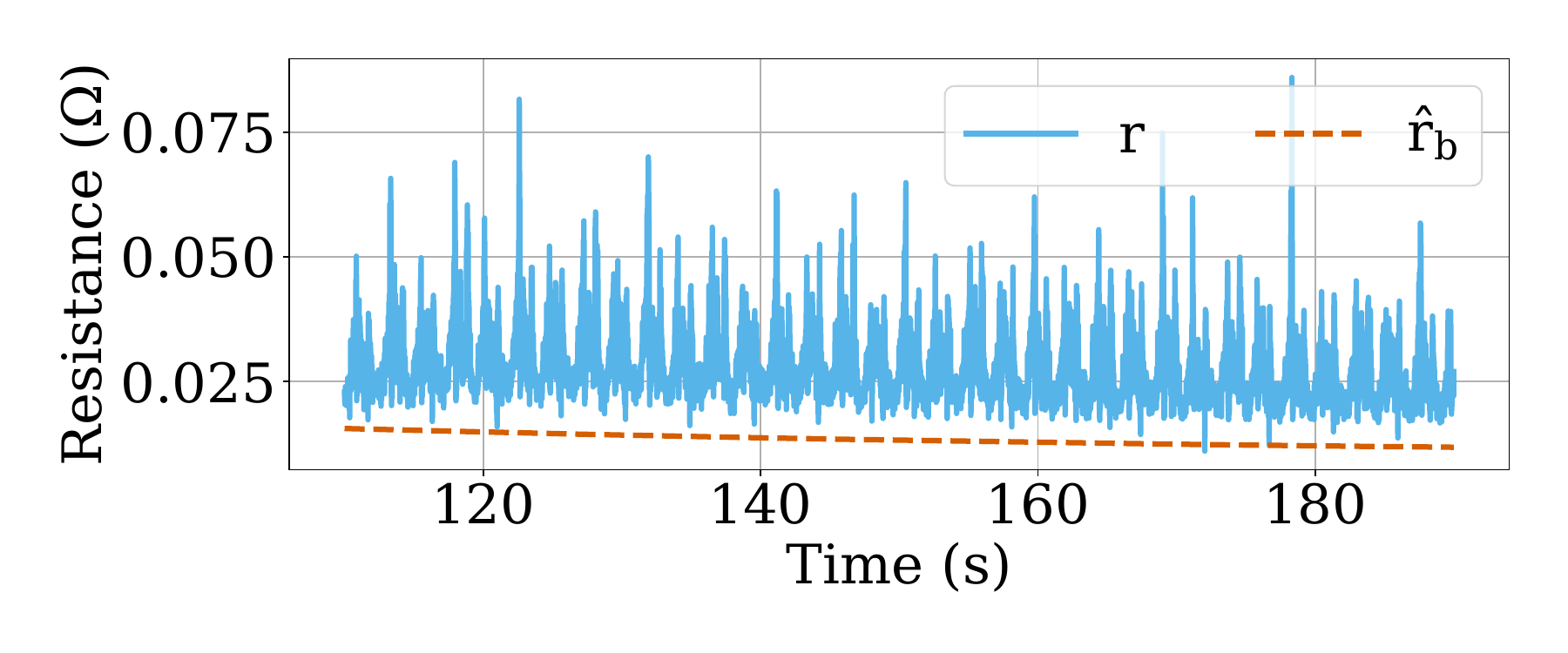}
        \caption{Comparison without temperature correction.}
        \label{fig:uncompensated_resistance}
    \end{subfigure}
    \begin{subfigure}[b]{\linewidth}
        \centering
        \includegraphics[width=\linewidth]{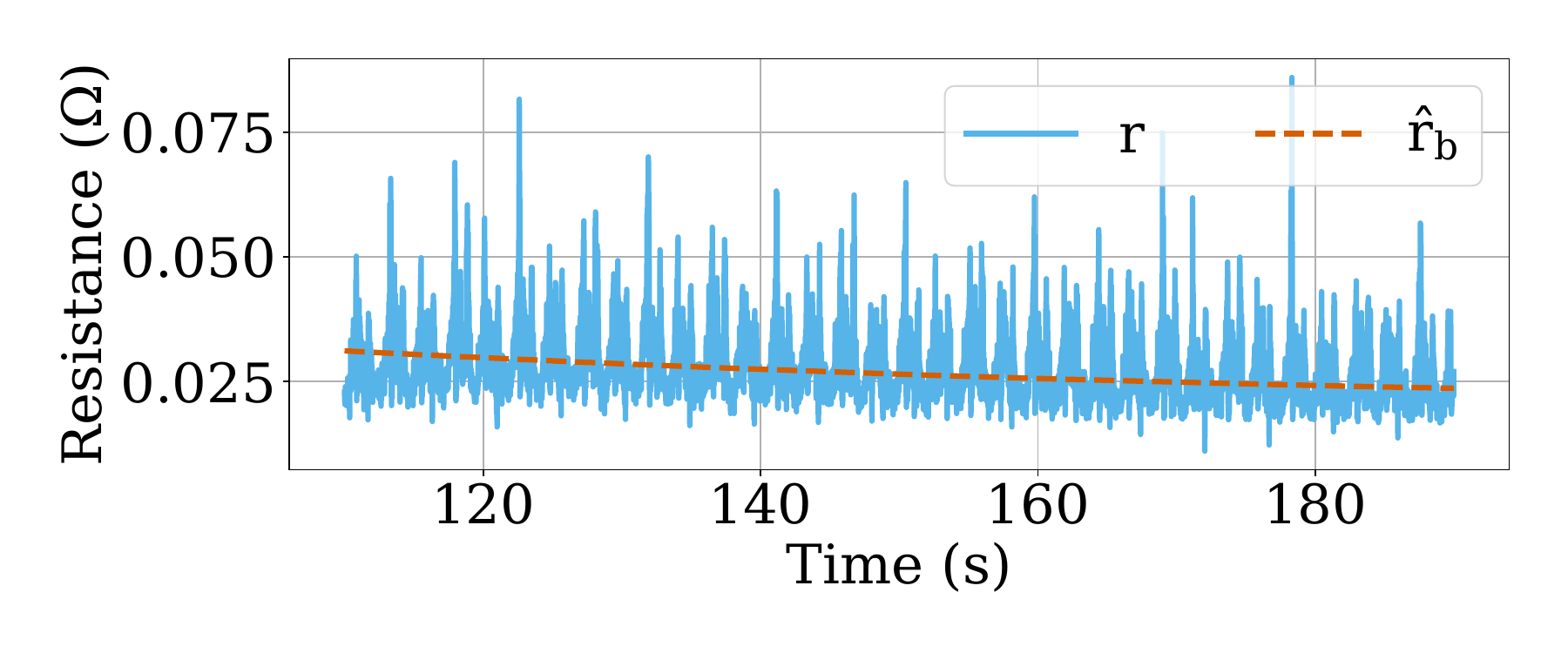}
        \caption{Comparison with temperature correction.}
        \label{fig:resistance}
    \end{subfigure}
    \caption{Comparison of estimated and measured internal resistance.}
    \label{fig:resistance_comparison}
\end{figure}

\begin{figure}
    \centering
    \includegraphics[width=\linewidth]{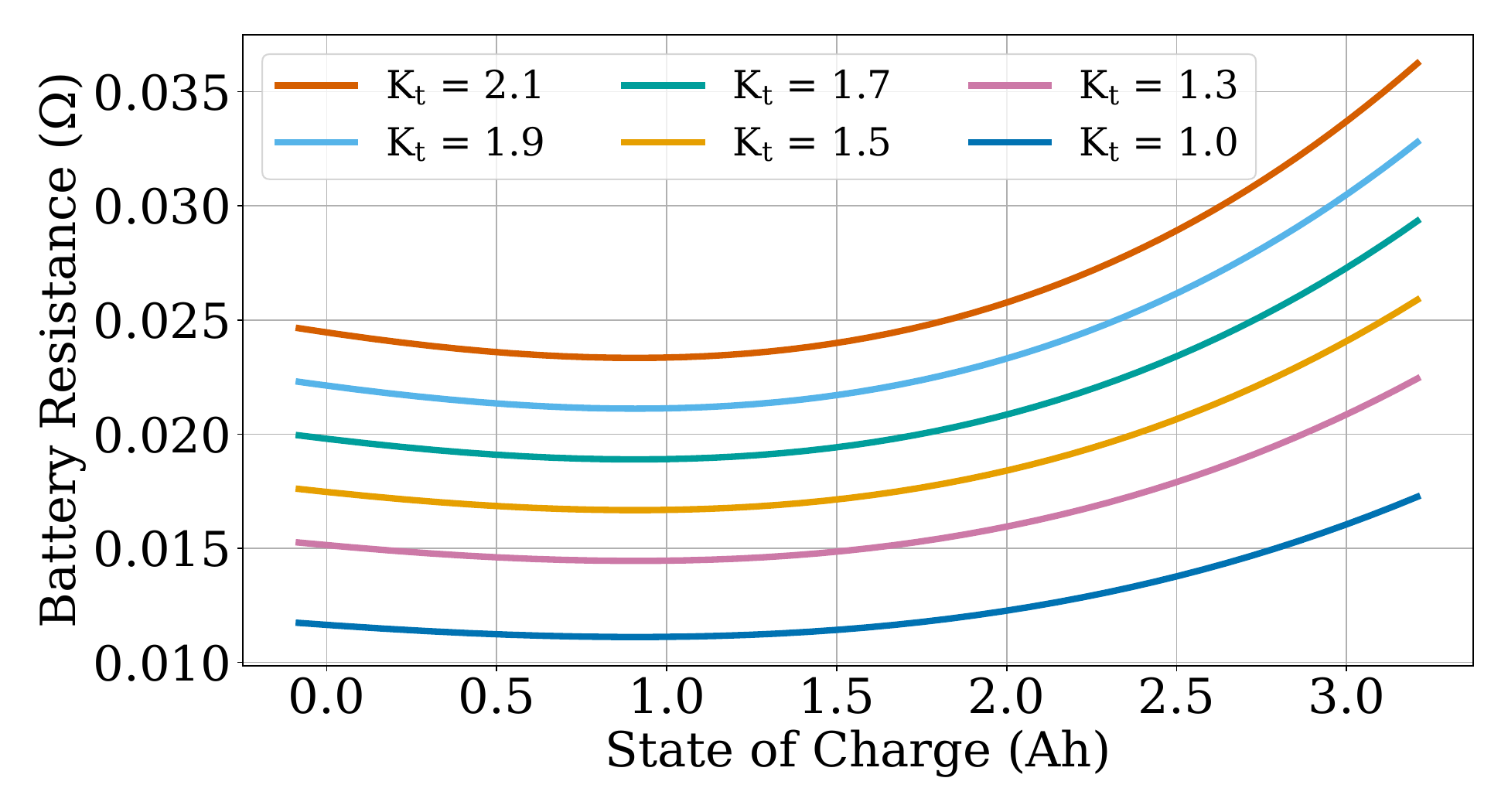}
    \caption{Projected scaling of internal resistance with temperature and \ac{soc} based on literature \cite{andreCharacterizationHighpowerLithiumion2011}.}
    \label{fig:resistance_temp_variation}
\end{figure}

We evaluate the accuracy of the predicted battery resistance $\estbatteryresistance$ by comparing with
\begin{equation}
\label{eq:estimated_resistance}
\measbatteryresistance = (\estbatteryvoltage - \meascircuitvoltage)~/~\meascircuitcurrent\text{.}
\end{equation}
It is to be noted that there is no direct measurement of the internal resistance, and it is therefore prone to noise propagated through $\meascircuitcurrent$, especially when the current drawn from the battery is low.
However, the estimated quantity $\estbatteryresistance$ did not match the indirect measurement $\measbatteryresistance$ in first validation flights, as shown in Figure \ref{fig:uncompensated_resistance}.
Battery testing was re-conducted and it was found that the discharge testing raised the battery's temperatures to nearly \SI{60}{\celsius}, whereas during the flight, the battery temperature remained a few degrees above ambient.
{
Based on the Temperature-\ac{soc}-Resistance relationships presented in literature \cite{andreCharacterizationHighpowerLithiumion2011}, it was concluded that the internal resistance has a large dependence on temperature and they form a 3D surface such that a scalar coefficient can be used to scale from a known resistance at a known temperature to a different temperature.
With this in mind, \eqref{eq:bat_res} was updated to include a temperature correction coefficient $\batterytempcoef$ such that
\begin{equation}
    \batteryresistance = \batterytempcoef(a_{\soctorescoef}\batterysoc^3 + b_{\soctorescoef}\batterysoc^2 + c_{\soctorescoef}\batterysoc + d_{\soctorescoef}) \label{eq:bat_res_temp} \text{.}
\end{equation}
An example of this scaling of internal resistance with temperature and \ac{soc} is shown in Figure \ref{fig:resistance_temp_variation}.
Even though the reference temperature value of \SI{60}{\celsius} was found during testing, a direct relationship between the actual temperature in flight and the temperature coefficient $\batterytempcoef$ cannot be established since commercial batteries with temperature sensing are not used in drone racing due to expensive and proprietary design of the battery required for sensing temperature between the cell packaging.
As such, our vehicle does not have a temperature sensor on the battery either, and so, a calibration flight was conducted before the real-world experiments where a best-fit scalar temperature correction coefficient $\batterytempcoef$ was calculated to compensate for the observed mismatch between $\estbatteryresistance$ and $\measbatteryresistance$.
In the future, an online Kalman observer could be used to estimate the temperature coefficient $\batterytempcoef$ during flight, and this could be used to update the internal resistance estimation.
}

\RCommon{This coefficient was then calculated through a short calibration flight in the real-world and was found to be $\batterytempcoef = 1.7$ for the data presented in Figure \ref{fig:resistance}.
Crucially, it was found that the temperature coefficient depended largely on the ambient weather conditions and once determined, it held throughout the experimental campaign for repeated flights, even with different batteries of the same specification.
The updated equation \eqref{eq:bat_res_temp} and the $\batterytempcoef$ coefficient value of $1.7$ were used for the simulation experiments as well as for the real-world analysis presented in the following sections.
}
The real-world validation experiments were then repeated after the calibration flight, and a comparison of the temperature-compensated $\measbatteryresistance$ and $\estbatteryresistance$ is shown in Figure \ref{fig:resistance}.
As seen in the plot, after the temperature calibration, the predicted resistance matches the average measured resistance during flight and confirms that the model can reliably predict the internal resistance of the battery during flight, which is crucial for accurate current and voltage predictions.

Building on top of the battery model, through the next sections, we present the validation of the predictions for circuit current ($\estcircuitcurrent$) and circuit voltage ($\estcircuitvoltage$).
Since $\meascircuitcurrent$ and $\meascircuitvoltage$ are directly measured quantities, the comparisons in the next section include the \ac{mae} and \ac{rmse} values of the estimation error.

\subsection{Current Estimation}\label{sec:current_estimation}
\begin{figure}
    \centering
    \includegraphics[width=\linewidth]{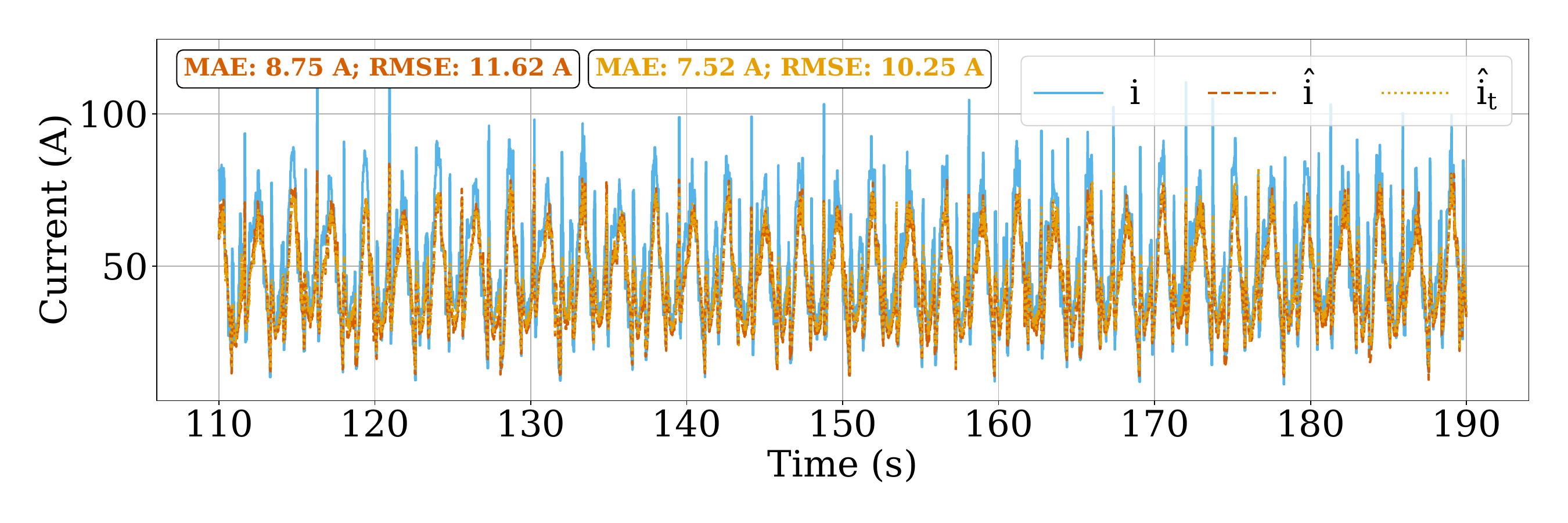}
    \caption{Estimated current from thrusts $\estcircuitcurrent$, estimated current from throttle $\estcircuitcurrent_t$, and measured current $\meascircuitcurrent$ drawn from the battery during agile flight.}
    \label{fig:current_agile}
\end{figure}

While $\estcircuitcurrent$ was already validated for $100\%$ throttle in Section \ref{sec:max_bench_verification}, in this section, we present additional validation of the model in a highly dynamic scenario.
During the flight, the \ac{nmpc} calculates the desired control input $\uavinput$ (=$\thrustvector$), which can, subsequently, be used to predict the current drawn during flight.
This predicted current $\estcircuitcurrent$ from the \ac{nmpc} is presented in comparison with the measured current $\meascircuitcurrent$ in Figure \ref{fig:current_agile}.
It is to be noted that the low-level controller of the \ac{uav} is responsible for converting the desired attitude rate to the desired motor throttles, and hence, the desired and applied throttles to the motors might differ and lead to a mismatch between the predicted and measured current.
The desired throttles from the low-level controller were also recorded and processed using \eqref{eq:throt_to_res} and \eqref{eq:volt_div} to obtain the predicted current $\estcircuitcurrent_t$.

The mean current for the flight shown in Figure \ref{fig:current_agile} was found to be \SI{50.71}{\ampere}.
For a mean prediction error of \SI{8.72}{\ampere} for $\estcircuitcurrent$, the error makes up approximately $17\%$ of the mean current, while $\estcircuitcurrent_t$ only makes up approximately $15\%$ of the mean current.
A lower error on $\estcircuitcurrent_t$ indicates the differences between desired throttles and applied throttles, arising from disturbance rejection performed by low-level controller.

The $17\%$ error is relatively moderate when considered against the high-agility and the dynamic nature of the flown trajectory.
However, this error can be attributed to the following factors:
\begin{itemize}
    \item the current consumption of the \ac{bldc} motor is dependent on the torque of the motor and since the tests were conducted on the static thrust stand with zero advance ratio, the dynamic torque on the motors from the high-velocity impingement of the airflow leads to non-zero advance ratio and therefore, higher demanded current \cite{coatesPropulsionSystemmodeling2019,moselerApplicationModelbasedFault2000}, and 
    \item the measurement of current draw through the onboard power module has non-linear scaling which caused higher current ranges to be more error-prone,
    \item the battery exhibits a transient capacitive effect wherein the resistance of the battery is significantly lower after a short period of recovery from a high-current draw.
\end{itemize}

However, we consider this prediction error to be relatively low in comparison to the peak current demand of $\SI{120}{\ampere}$ in the flight, and it was shown to be sufficient and accurate for the purpose of predicting the available power and thrust limits for the \ac{nmpc}, as shown in Section \ref{sec:max_bench_verification}.

\subsection{Terminal Voltage Estimation}
\begin{figure}
    \centering
    \includegraphics[width=\linewidth]{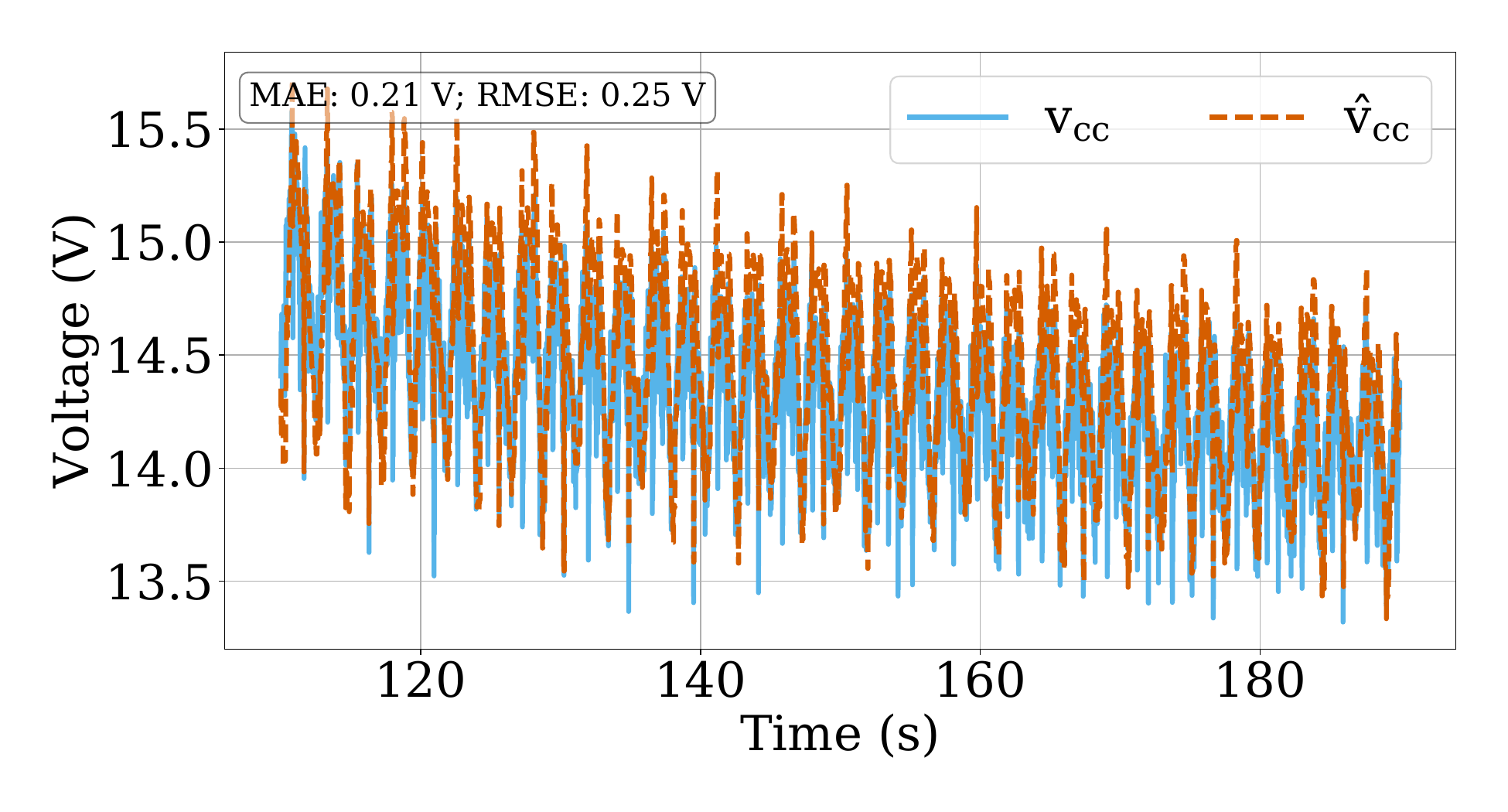}
    \caption{Terminal voltage estimation.}
    \label{fig:voltage_agile}
\end{figure}

Building on the estimated quantities of $\estbatteryvoltage, \estcircuitcurrent, \estbatteryresistance$, we can estimate the terminal voltage $\estcircuitvoltage$ as 
\begin{equation}
\label{V_2}
\estcircuitvoltage = \estbatteryvoltage - \estcircuitcurrent\estbatteryresistance\text{.}
\end{equation}
While $\estcircuitvoltage$ does not serve as a directly used quantity in our proposed work, it has several applications in \ac{uav} flights such as
\begin{itemize}
    \item confirmation for mission termination if terminal voltage is estimated to fall below the minimum operating voltage of sensors and onboard electronics, and, 
    \item enable predictive throttle value reference to low-level controller to account for the subsequent voltage drop.
\end{itemize}
We present a comparison of the $\estcircuitvoltage$ with measured terminal voltage $\meascircuitvoltage$ in Figure \ref{fig:voltage_agile}, along with \ac{mae} and \ac{rmse} values.

Due to the accumulation of error from other estimated quantities, a moderate error is observed in the comparison.
However, even after stacking of several predictions, the voltage predictions prove to be a healthy indicator of the model's fit and predictive abilities.

\subsection{Computation Time Analysis}
\begin{figure}
    \centering
    \includegraphics[width=\linewidth]{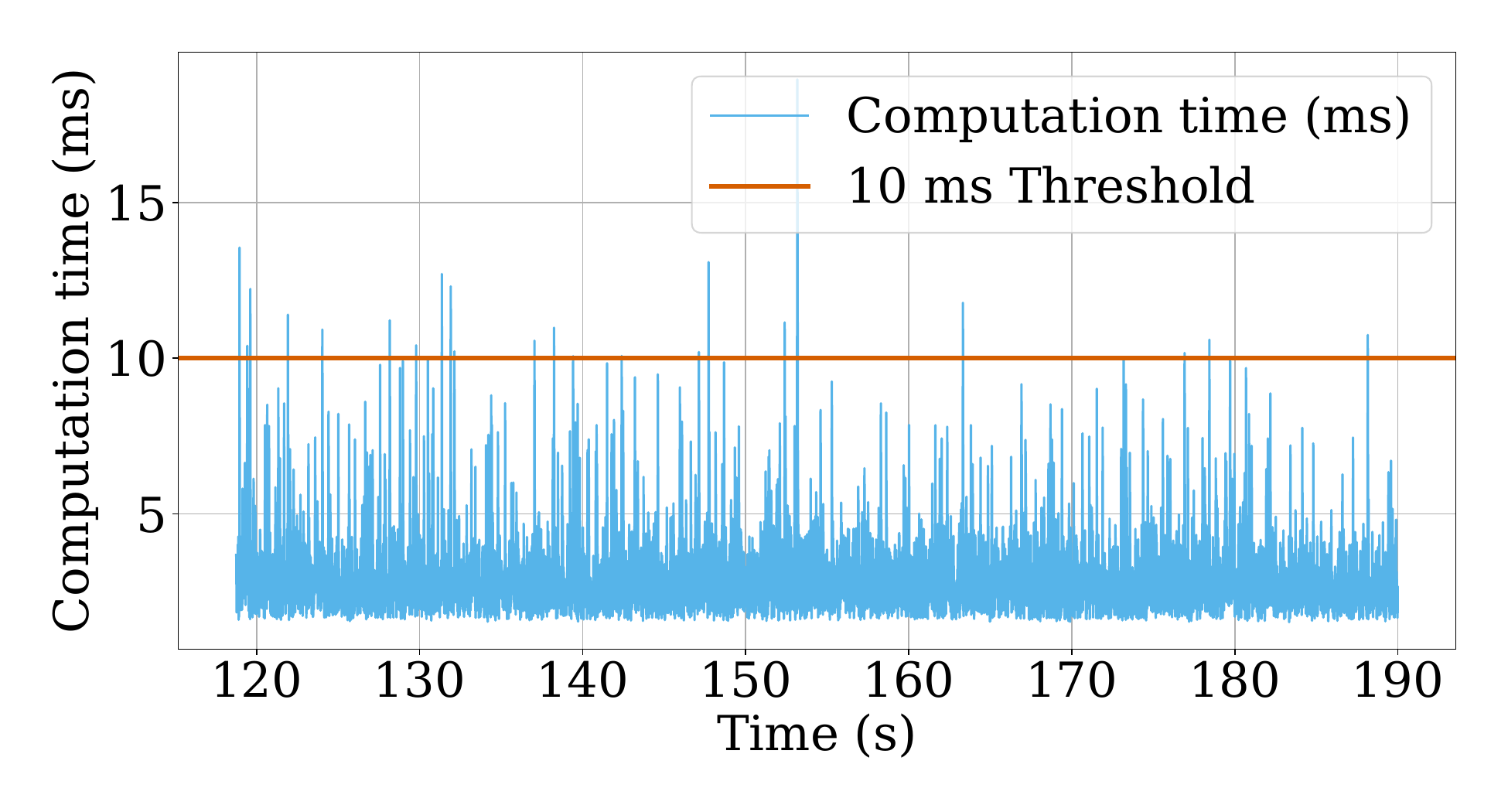}  
    \caption{Computation time of the controller during flight for every run of the NMPC.}
    \label{fig:comp_time}
\end{figure}

A significant portion of work in the literature is dedicated to simplification of models for efficient computation on \acp{uav} and this has been a limiting factor for general deployment of complex models.
We use polynomial methods in this work to achieve the desired computational efficiency and as such, it needs to be verified in real-life examples of demanding trajectories.
The plot for computation time during real-world flight for each \ac{nmpc} cycle can be seen in Figure \ref{fig:comp_time}. 
Upon analysis, it was found that it healthily meets the computational time limit of \SI{10}{\milli\second}.
\RThree{In fact, the mean computation time was found to be only \SI{2.85}{\milli\second}, with a maximum computation time of \SI{18.10}{\milli\second}, and a standard deviation of \SI{1.25}{\milli\second}.
Out of the $7920$ runs of the \ac{nmpc} in flight, violation of computation time was limited to $33$ instances ($\SI{0.42}{\percent}$) even in a challenging trajectory over the course of $\approx\SI{100}{\second}$ flight.
In case of violation, the \ac{nmpc} simply skipped to the next state update, and since there were so few instances of violation, no adverse effects on the flight performance were observed.}

\subsection{Results and Discussion}
We proposed this model to bridge the gap between complex offline models and simplified real-time models through polynomial functions that can model each aspect of the propulsion system.
Through our extensive, rigorous, and challenging real-world experiments, we have confirmed the computation viability and predictive capabilities of our proposed `BC-NMPC` model.
Since the current consumption and power consumption are dependent only on the \ac{nmpc} control inputs, the model offers effective predictions of energy, power, and flight time for payload transport even in cases of wind disturbance.
Additionally, for drone racing scenarios, the ability to predict the maximum available collective thrust can be used to replan demanding trajectories to keep the \ac{uav} at its limits and achieve time-optimal flight.
We demonstrate this ability and its advantages in the following section.

\section{Simulation experiment}
During a usual planned mission or flight, the \ac{uav} follows a pre-determined feasible trajectory expressed as a set of references $\statestddesired$ as described in \eqref{eq:nonlinear_constraints}.
\RCommon{In this scenario, two primary problems appear due to the nature of trajectory tracking at high-speed and high-acceleration, and we propose a solution for each of these problems.

Firstly, even if the trajectory is planned under the constraint of a certain peak thrust, the \ac{nmpc} deals with tracking errors by treating them as disturbances, and in doing so, needs to utilise more thrust than what was planned in the trajectory.
For any given \ac{uav} flying close to its limits, reduction of the tracking error is contingent on the thrust headroom available to its controller.
When the thrust headroom shrinks as the battery discharges, an unaware controller asks for a higher thrust than available and continuously misses the acceleration targets required for tracking error reduction.
We propose that an aware controller can judiciously plan for the gradual reduction in the thrust to utilise feasible acceleration targets more effectively.

Secondly, for a \ac{uav} with shrinking thrust limits, there are usually two options for planning trajectories.
When planned with optimality, the trajectory can assume static thrust limits from a fully-charged battery \cite{cpc}, and therefore, when the available thrust depletes, the control input calculated by \ac{nmpc} might not be feasible, and thus, lead to deviation, or a collision.
Conversely, the trajectory can be planned conservatively by setting the static thrust limit at the lowest available thrust in the discharged state to forego time optimality in favour of safe operation.
We propose a third alternative in the form of a trajectory replanning approach where the trajectory planner continuously receives the maximum available acceleration directly from the controller, and replans the trajectory throughout the flight.
This allows the \ac{uav} to stay near the available thrust limit throughout the flight and maintain time optimality for the entire duration.
We present and study these two proposals in the form of an ablation study in a simulation environment using the MRS UAV system~\cite{Baca2021}.
The simulation utilised a digital twin of the \ac{uav} but with one key distinction: the mass of the \ac{uav} was kept at its standard of \SI{1.54}{\kilo\gram} to highlight the \ac{uav} performance degradation under decreasing thrust limits.
The battery was simulated with the same characteristics as the real-world twin (\ac{lipo}) and the experiment was terminated once the battery discharged to a $\batteryvoltage = \SI{14.4}{\volt}$ or $\SI{3.6}{\volt}$ per cell.
The battery was set to emulate the $\batterytempcoef = 1.7$ for the simulation, leading to a $\collectivethrust_\text{max}=\SI{56}{\newton}$ at full charge and a $\collectivethrust_\text{max}=\SI{49}{\newton}$ at $\SI{14.4}{\volt}$.
The ablation study is presented through four experiments, each repeated thrice for confirmation.
The experimental results are presented in their respective sections, and discussed in detail in sections \ref{sec:thrust_awareness_discussion} and \ref{sec:time_optimality_discussion}.

\paragraph{Unaware} In this experiment, the trajectory was pre-planned to be executed at $\SI{3.5}{\accelgrav}$ of acceleration which set the \ac{rtt} at $\SI{52.8}{\newton}$ for a \ac{uav} mass of $\SI{1.54}{\kilo\gram}$.
Since the $\collectivethrust_\text{max}$ was $\approx \SI{56}{\newton}$ at the beginning of the flight, the \ac{uav} had reserve thrust for torque allocation inside the low-level controller until the \ac{rtt}.
The controller was kept unaware of the time-varying thrust limit, but the thrust limit was simulated as a cutoff to simulate the effects of battery discharge mid-flight.
In this setup, it was expected that the \ac{uav} would accrue tracking error once the available thrust dropped below the \ac{rtt}, and in order to overcome disturbances, the controller would saturate the thrust beyond $\collectivethrust_\text{max}$ and up to the theoretical limit of the actuators.
The plots of this experiment are shown in Figure \ref{fig:unaware_combined}.

\begin{figure}
\centering
\begin{subfigure}[b]{\linewidth}
\centering
\includegraphics[width=\linewidth]{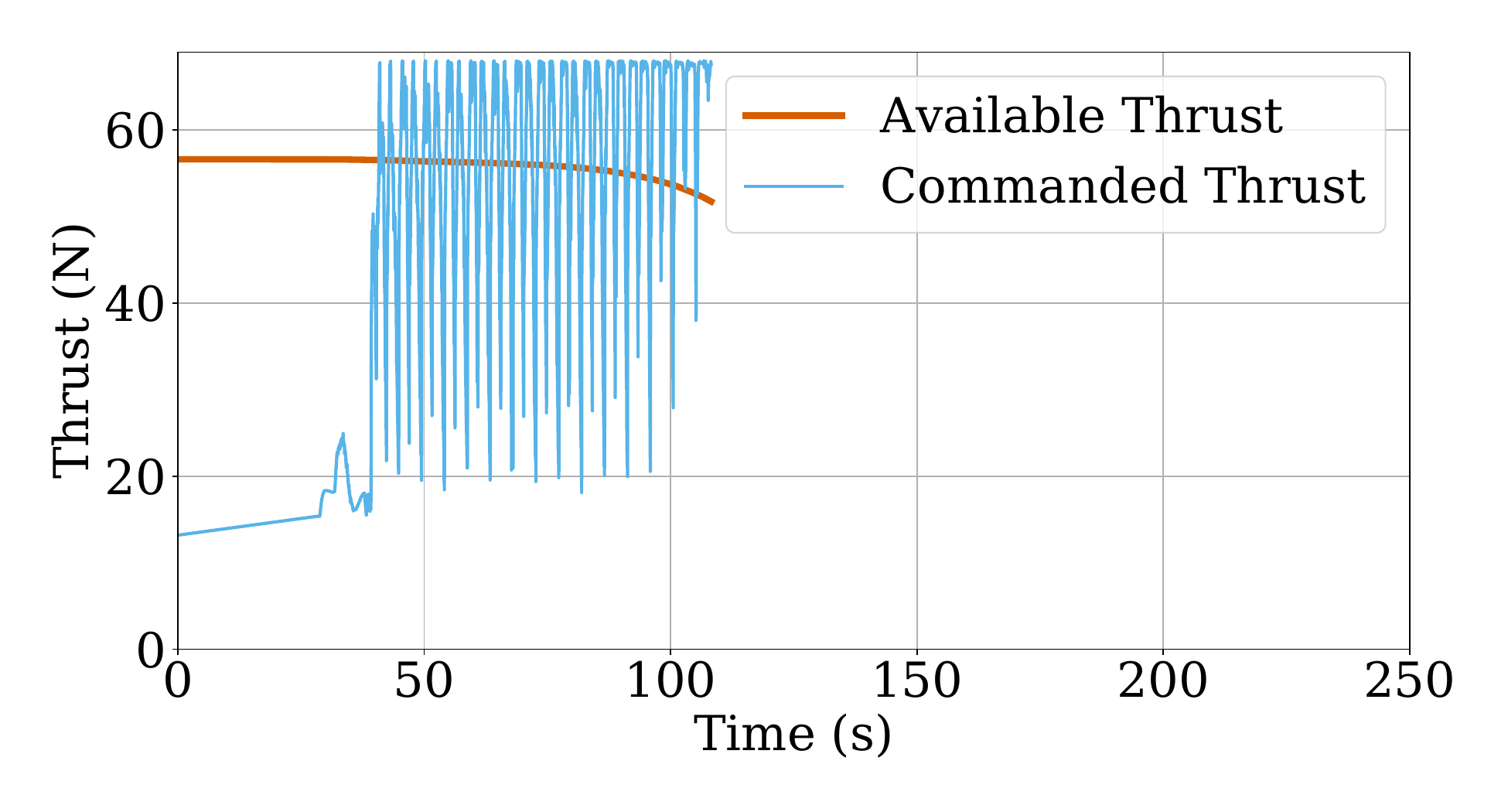}
\caption{Commanded thrust vs. available thrust during flight.}
\label{fig:unaware_thrust}
\end{subfigure}
\begin{subfigure}[b]{\linewidth}
\centering
\includegraphics[width=\linewidth]{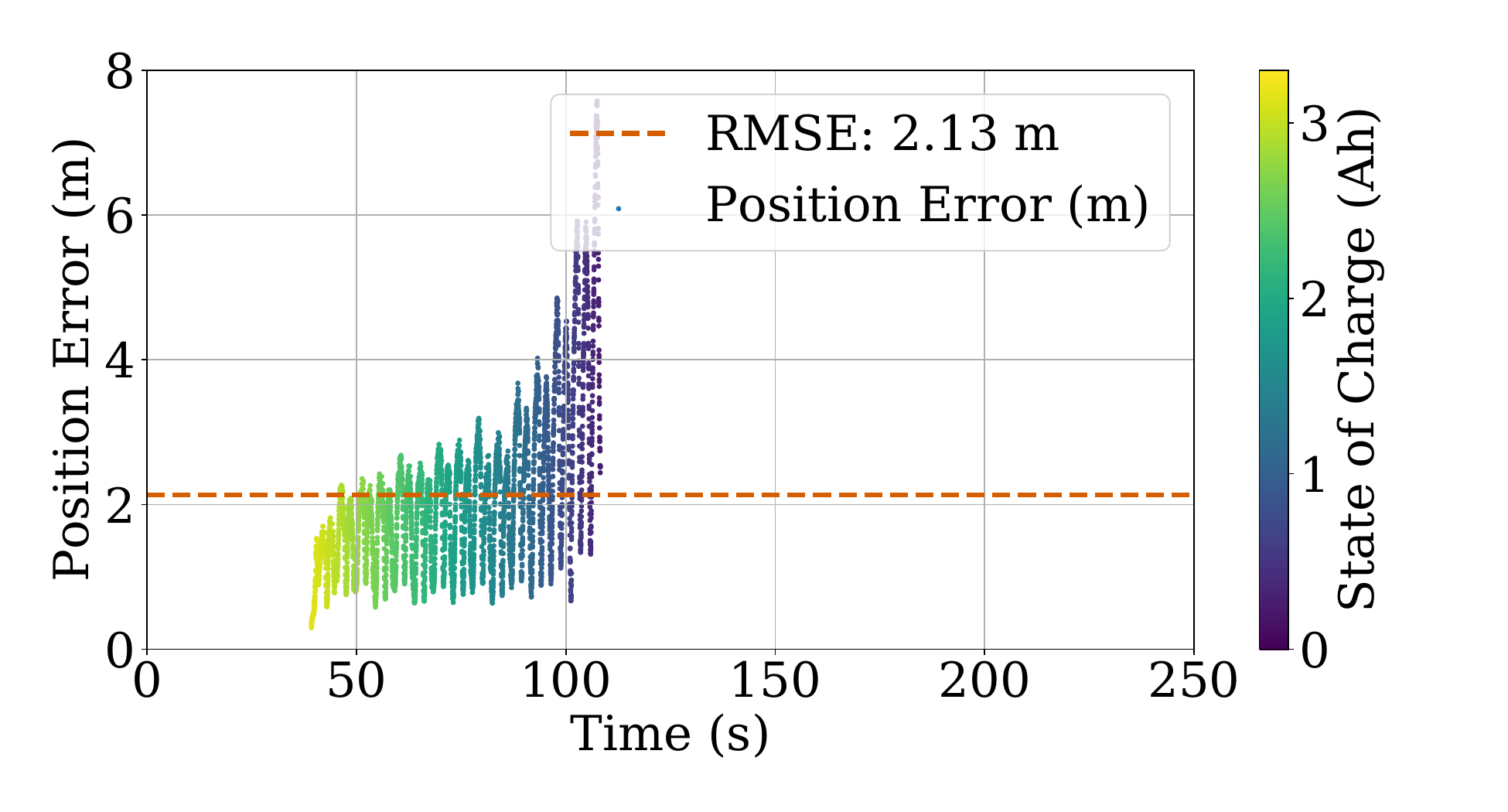}
\caption{Position error during flight.}
\label{fig:unaware_pos_error}
\end{subfigure}
\begin{subfigure}[b]{\linewidth}
    \centering
    \includegraphics[width=\linewidth]{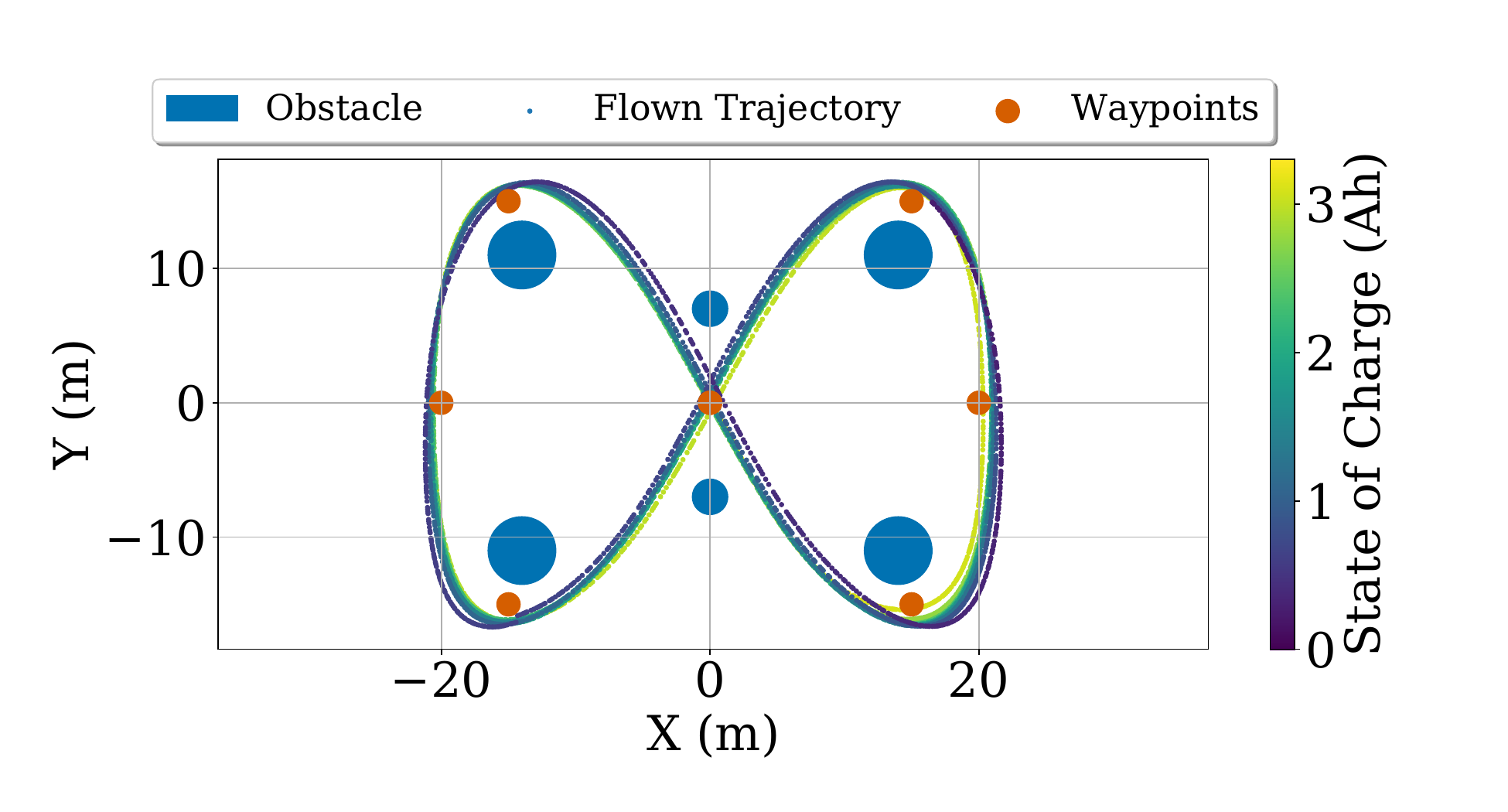}
    \caption{Trajectory of flight.}
    \label{fig:unaware_trajectory}
\end{subfigure}
\caption{Thrust, position error, and the trajectory of the `Unaware` flight.}
\label{fig:unaware_combined}
\end{figure}

\paragraph{Aware} In this experiment, the trajectory and the simulation setup remained the same as the 'Unaware' experiment described above, but the controller included the formulation described in Section \ref{sec:bc_nmpc_description}.
In this setup, it was expected that the \ac{uav} would accrue the tracking error slower than the `Unaware` experimental setup, even as the controller was expected to abide by the shrinking $\collectivethrustmax$.
The plots of this experiment are shown in Figure \ref{fig:aware_combined}.

\begin{figure}
    \centering
    \begin{subfigure}[b]{\linewidth}
        \centering
        \includegraphics[width=\linewidth]{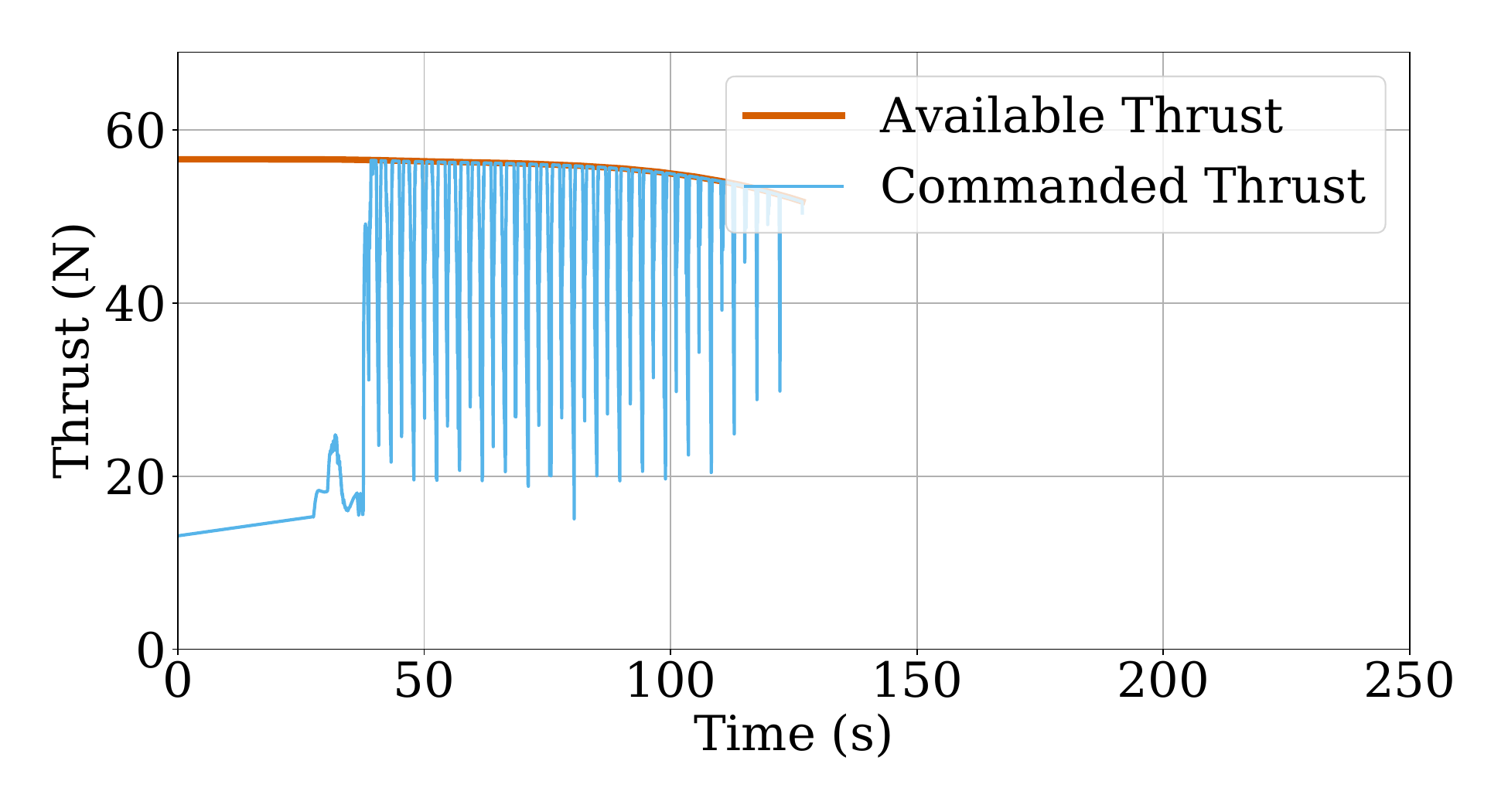}
        \caption{Commanded thrust vs. available thrust during flight.}
        \label{fig:aware_thrust}
    \end{subfigure}
    \begin{subfigure}[b]{\linewidth}
        \centering
        \includegraphics[width=\linewidth]{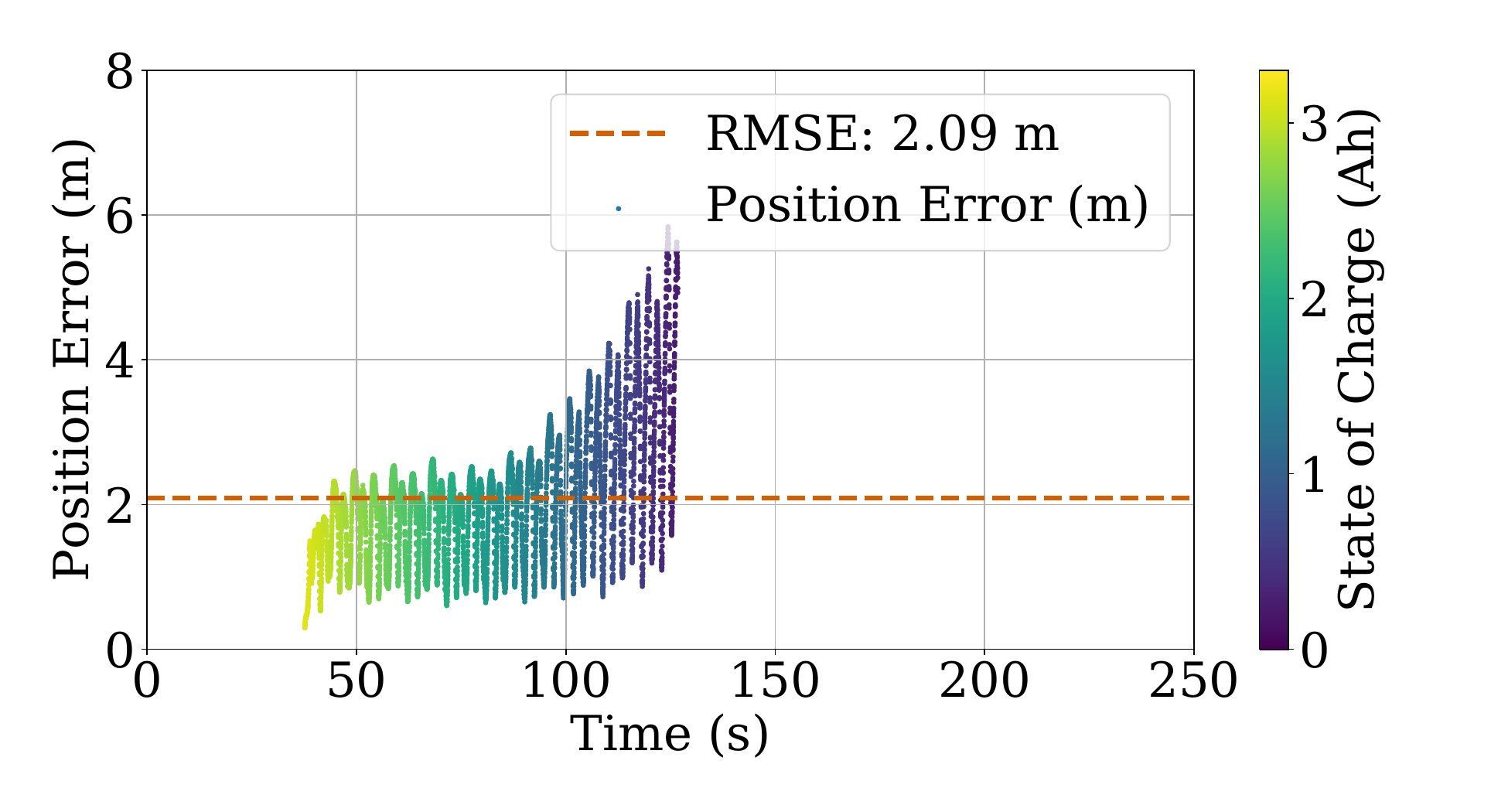}
        \caption{Position error during flight.}
        \label{fig:aware_pos_error}
    \end{subfigure}
    \begin{subfigure}[b]{\linewidth}
        \centering
        \includegraphics[width=\linewidth]{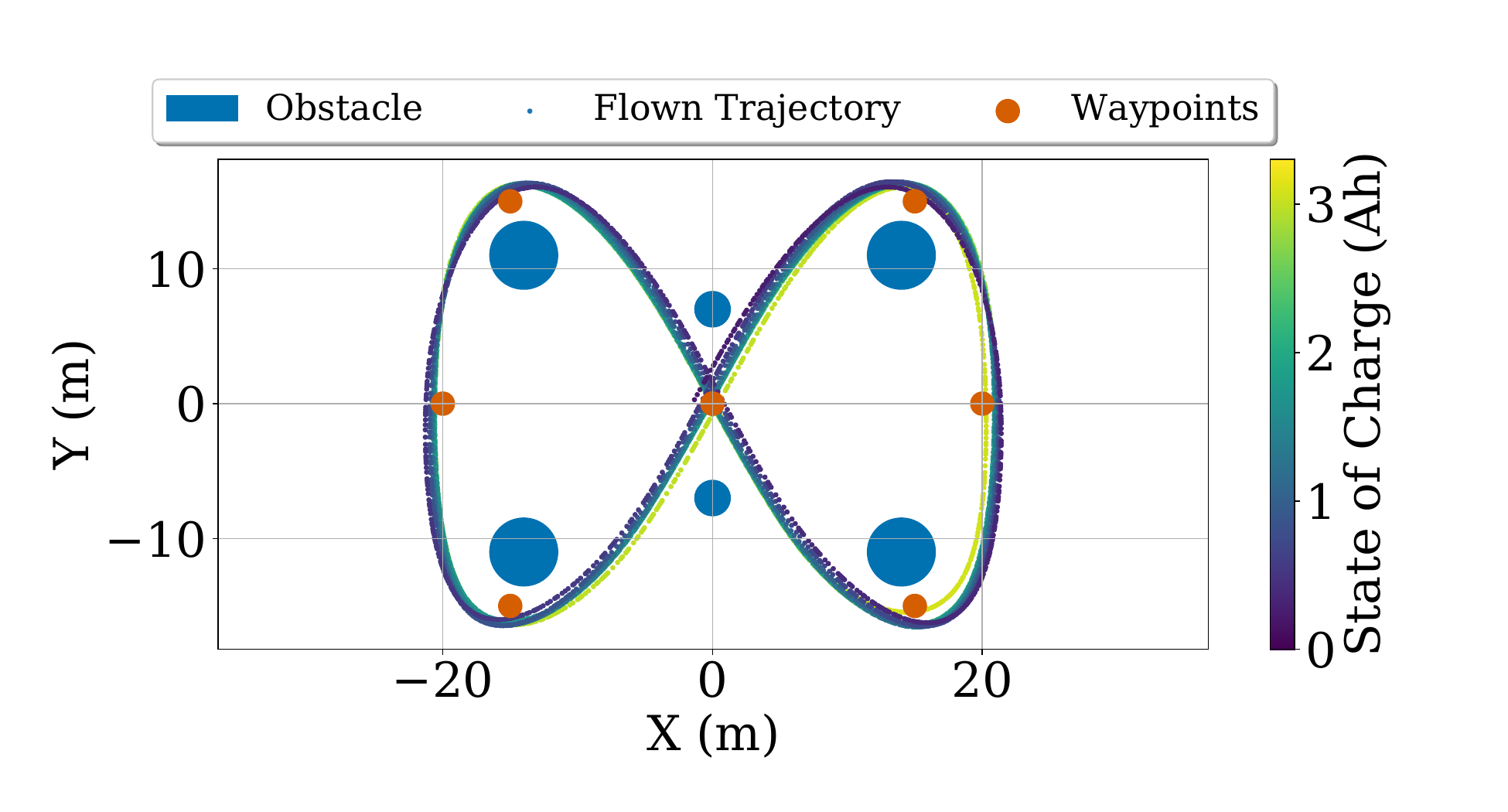}
        \caption{Trajectory of flight.}
        \label{fig:aware_trajectory}
    \end{subfigure}
    \caption{Thrust, position error, and the trajectory of the 'Aware' experiment.}
    \label{fig:aware_combined}
\end{figure}

\paragraph{Conservative} In this experiment, similar to the `Unaware` experiment, the controller was kept unaware of the shrinking thrust limits, but the trajectory was planned conservatively with a $\collectivethrustmax = \SI{49}{\newton}$ so as to remain within the thrust limit at all times.
It was expected that the mean position error would remain nearly constant throughout the flight, but the controller's desired peak thrust would breach the $\collectivethrustmax$ to overcome position tracking errors.
The plots of this experiment are shown in Figure \ref{fig:conservative_combined}.

\begin{figure}
\centering
\begin{subfigure}[b]{\linewidth}
\centering
\includegraphics[width=\linewidth]{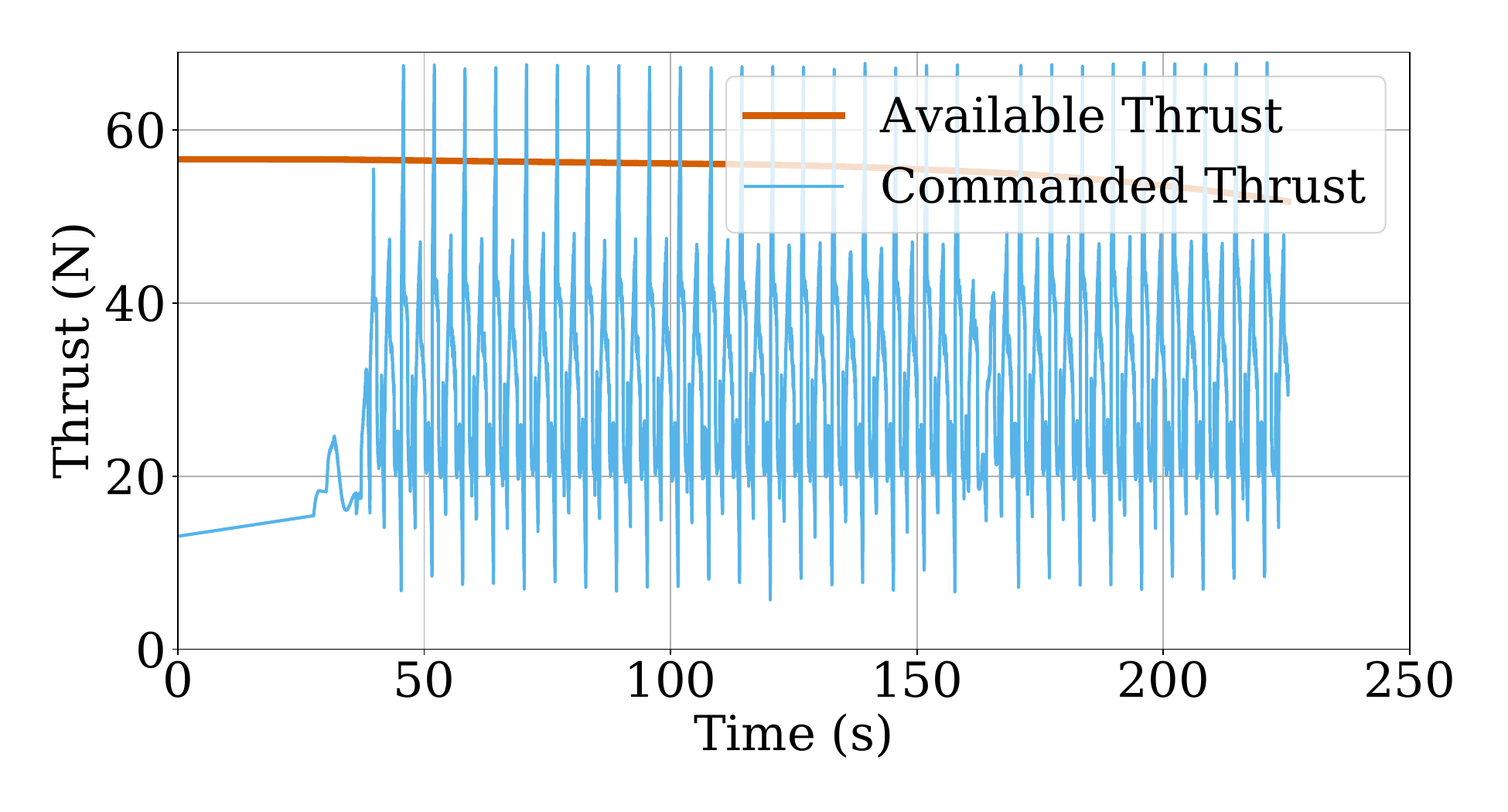}
\caption{Commanded thrust vs. available thrust during flight.}
\label{fig:conservative_thrust}
\end{subfigure}
\begin{subfigure}[b]{\linewidth}
\centering
\includegraphics[width=\linewidth]{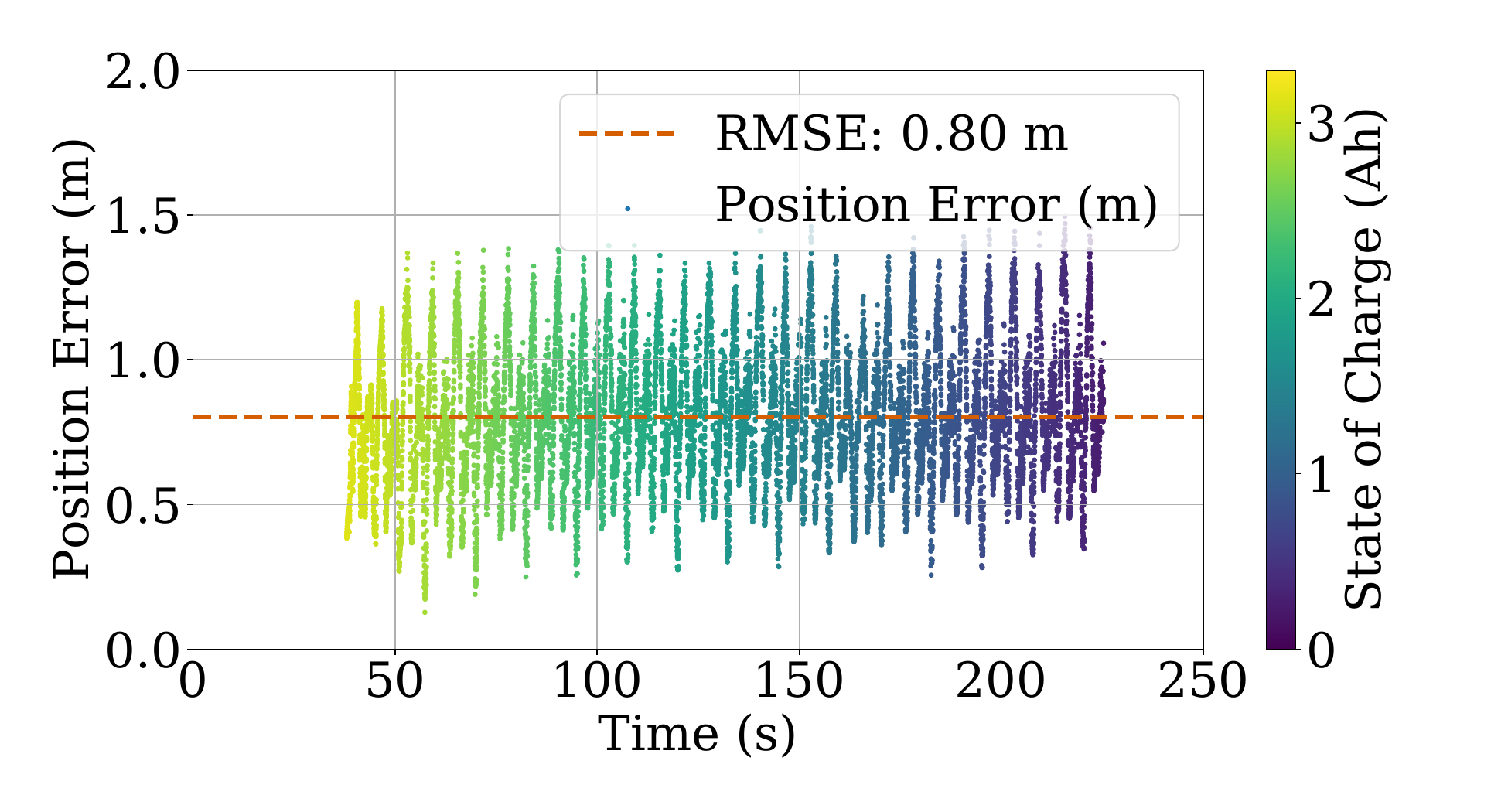}
\caption{Position error during flight.}
\label{fig:conservative_pos_error}
\end{subfigure}
\begin{subfigure}[b]{\linewidth}
    \centering
    \includegraphics[width=\linewidth]{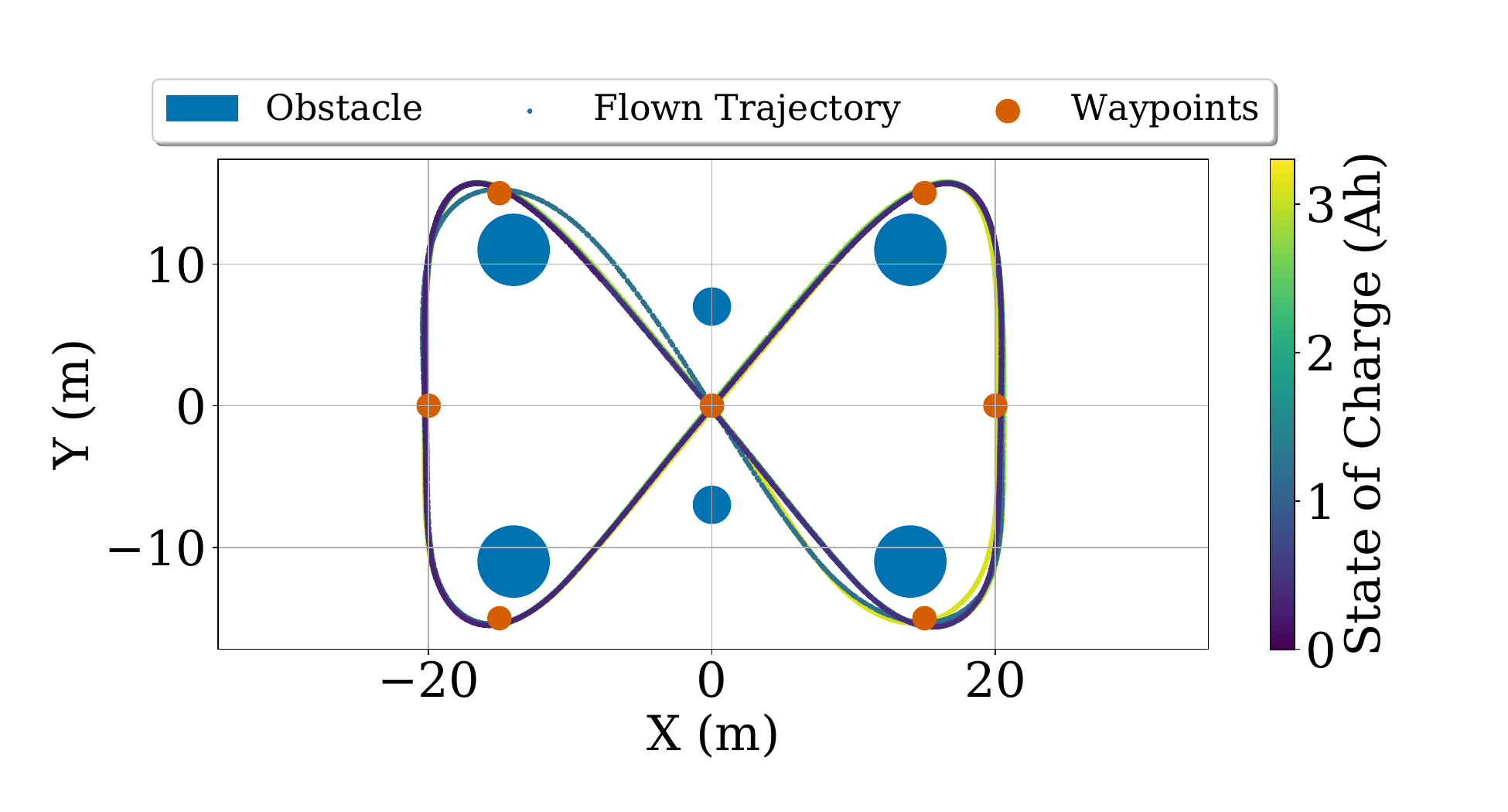}
    \caption{Trajectory of flight.}
    \label{fig:conservative_trajectory}
\end{subfigure}
\caption{Thrust, position error, and the trajectory of the 'Conservative' experiment.}
\label{fig:conservative_combined}
\end{figure}

\paragraph{Replanned} Similar to the 'Aware' experiment, the controller was aware of the shrinking thrust limit in this experiment.
Crucially, unlike the 'Aware' experiment, the trajectory was replanned online using the PMM planner \cite{PMM} with the same waypoints as previous experiments but with acceleration limits informed by the controller.
This planner was chosen for its real-time computability and the ability to generate thrust limited time-optimal trajectories given a specific available collective thrust.
Each time the \ac{uav} arrived within $\SI{2}{\metre}$ radius of the waypoint, the remaining trajectory was replanned using the new available acceleration.
In this way, the time optimality could be maintained throughout the flight, and it was expected that the controller would stay within the $\collectivethrustmax$ limit while flying faster than the `Conservative` approach.
The plots of this experiment are shown in Figure \ref{fig:replanned_combined}.\\

\begin{figure}
\centering
\begin{subfigure}{\linewidth}
\centering
\includegraphics[width=\linewidth]{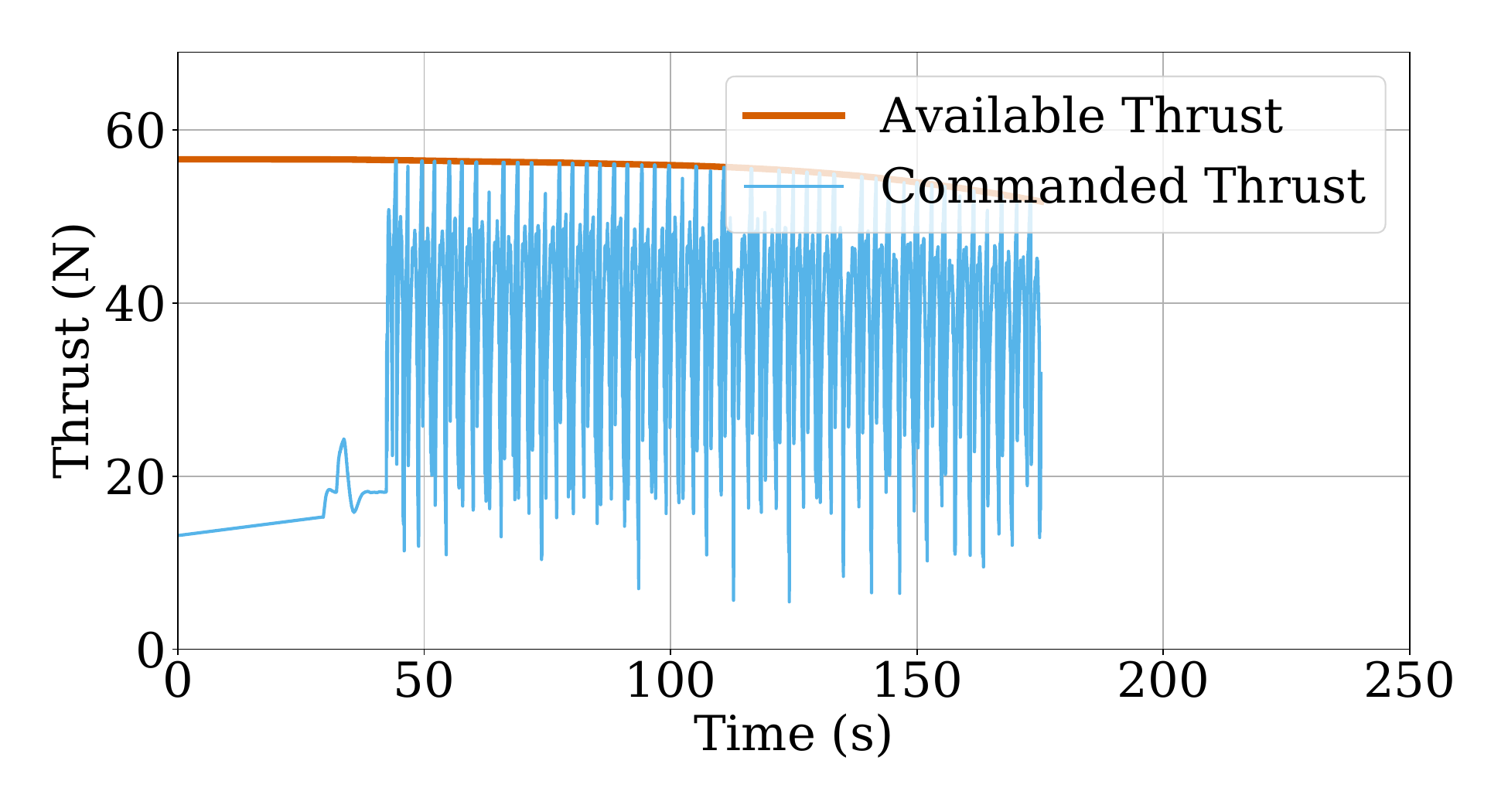}
\caption{Commanded thrust vs. available thrust during flight.}
\label{fig:replanned_thrust}
\end{subfigure}
\begin{subfigure}{\linewidth}
\centering
\includegraphics[width=\linewidth]{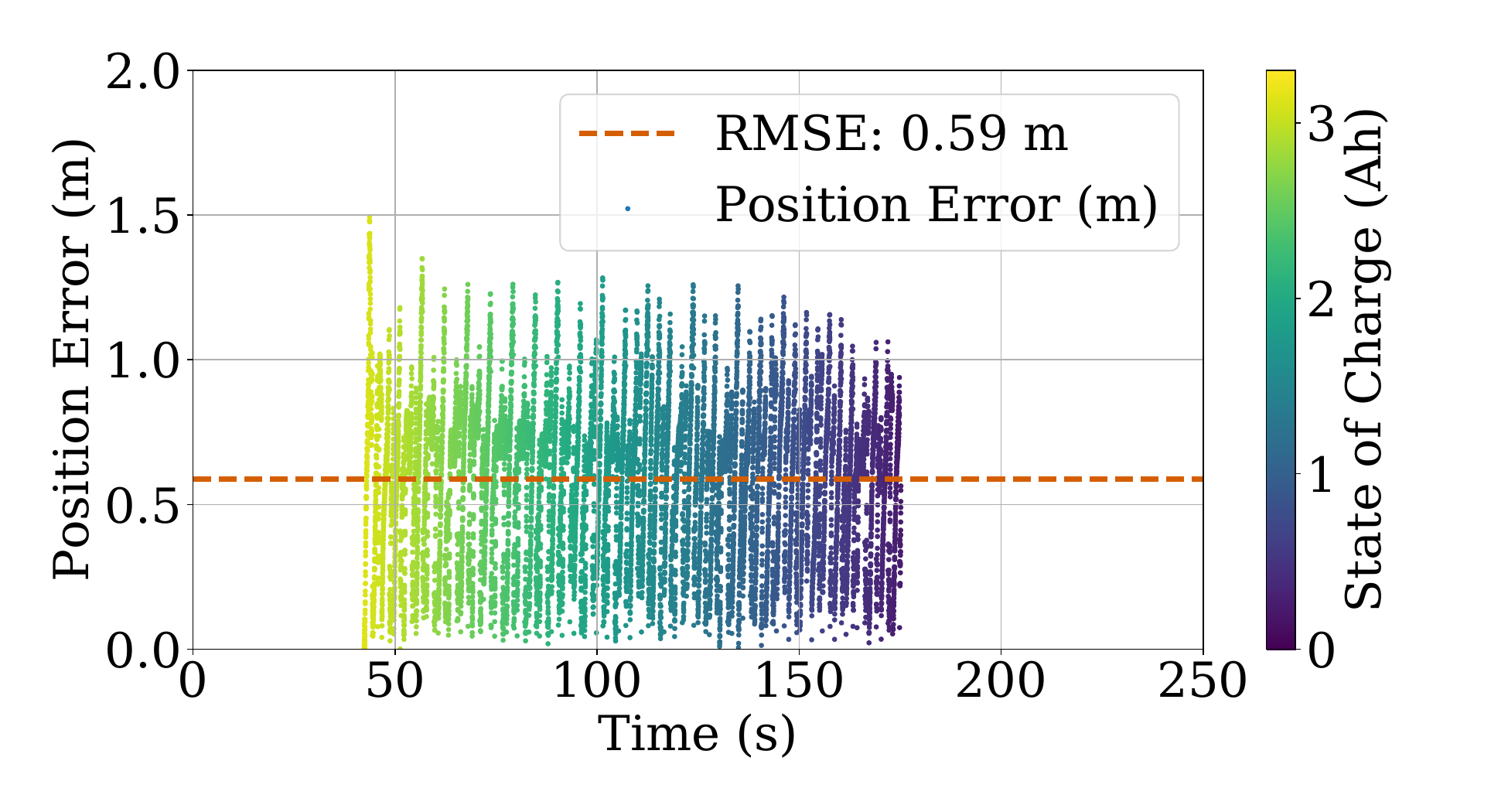}
\caption{Position error during flight.}
\label{fig:replanned_pos_error}
\end{subfigure}
\begin{subfigure}{\linewidth}
    \centering
    \includegraphics[width=\linewidth]{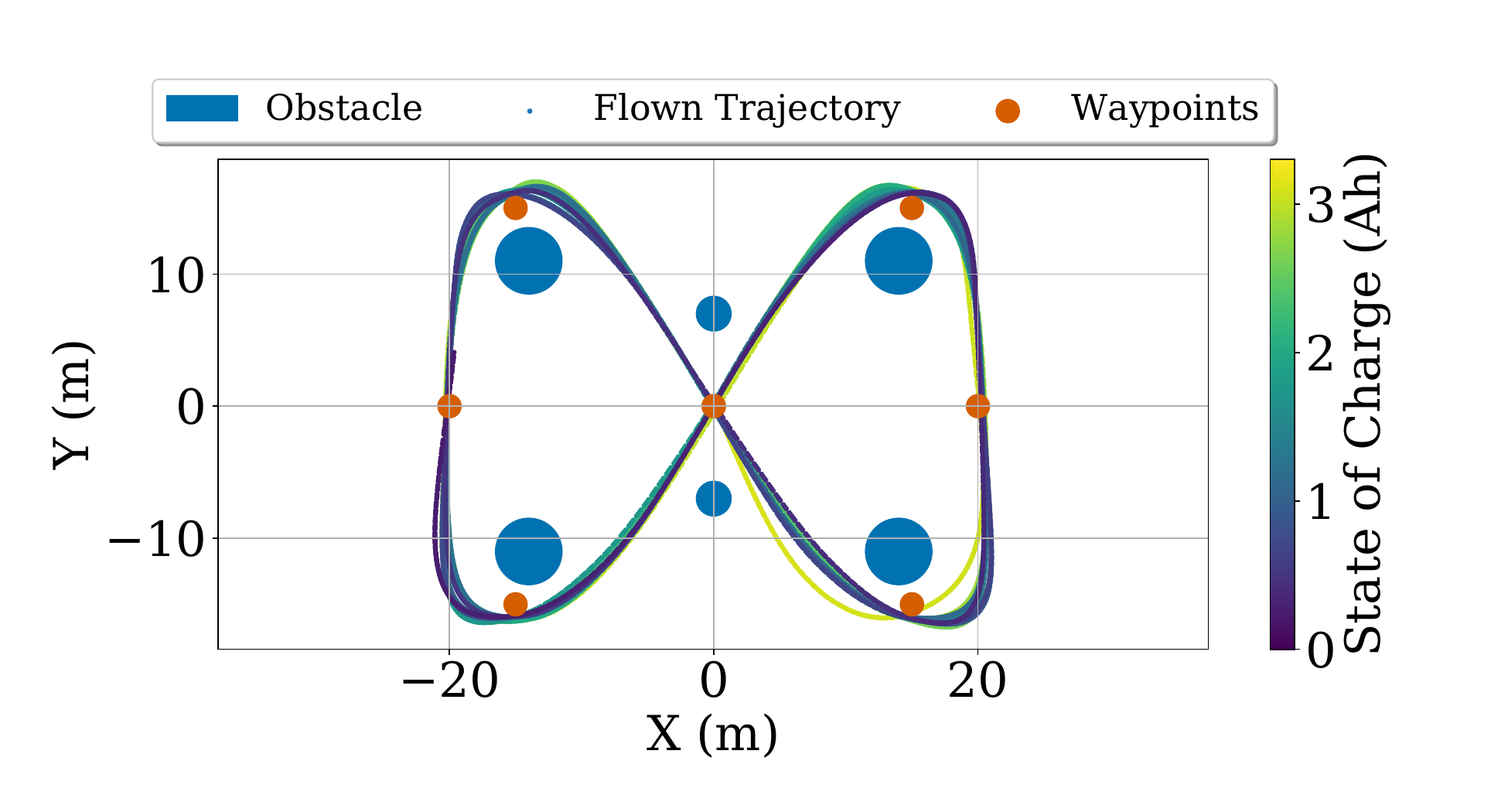}
    \caption{Trajectory of flight.}
    \label{fig:replanned_trajectory}
\end{subfigure}
\caption{Thrust, position error, and the trajectory of 'Replanned' experiment.}
\label{fig:replanned_combined}
\end{figure}

Figure \ref{fig:unaware_combined}, Figure \ref{fig:aware_combined}, Figure \ref{fig:conservative_combined}, and Figure \ref{fig:replanned_combined} show the plots for collective thrust, position error (or 3D tracking error) from the reference given to the \ac{nmpc}, and the overhead view of the trajectory flown.
In addition to these figures, Figure \ref{fig:soc_vs_performance} presents the key comparison metrics between all four approaches in a single place for easier understanding.

\subsection{Importance of Thrust-Awareness in Flight}
\label{sec:thrust_awareness_discussion}

For high-speed flight, a large emphasis is placed on modelling accuracy inside the \ac{nmpc} because multi-objective optimisation can more effectively make trade-offs when the demanded performance lies outside physical limits of the system.
In line with these expectations, Figure \ref{fig:soc_vs_performance} clearly indicates the large benefits of flying with the awareness of the physical limits of the aircraft.
The 'Aware' controller was able to reliably predict the trade-off of maximum available thrust to judiciously apply thrust and torque inputs, and this led to an improvement in both the mean position error and the peak position error. 
The mean position error diverged significantly between the 'Unaware' and the 'Aware' approaches at the end of the \ac{soc}, and the `Aware` controller showed a significant reduction of $\SI{24.9}{\percent}$ at the final juncture.
The rate of rise of the mean position error is also much slower for the `Aware` controller, as predicted, and the explanation becomes quite clear from Figures \ref{fig:unaware_thrust} and \ref{fig:aware_thrust} as the `Unaware` controller continues to saturate its thrust command beyond the physical capability of the aircraft.
This observation clearly highlights the role of reserve thrust for a multi-rotor vehicle flight, since, a saturation of thrust in high-level controller leads to deterioration of torque allocation in the low-level controller.
As the maximum thrust depleted, the \ac{uav} struggled to apply enough torque for tracking the curved trajectory, and the position error rose once the predicted thrust limit dropped below the required thrust for the trajectory. 
Thrust awareness also had a noticeable impact on the peak position error as shown in Figure \ref{fig:soc_vs_peak_rmse}, where the 'Aware' controller showed a significant reduction of $\SI{22.9}{\percent}$ at the end.
In terms of velocity over the flight duration, Figure \ref{fig:soc_vs_velocity_fit} shows a nearly matched velocity profile owing to both the controllers flying the same trajectory.
While it was not the objective of this work, duration of the flight also increased slightly for the `Aware` controller due to lesser power consumption arising from significantly lower thrust budget.

Through these results, we confirm that the proposed modelling approach brings significant reductions in the position error, even in the absence of a replanner, for high-acceleration and high-velocity flights.

\subsection{Time optimal planning vs. Conservative planning}
\label{sec:time_optimality_discussion}
As described earlier, the 'Conservative' approach sacrificed time optimality for a safer flight and as expected, Figure \ref{fig:soc_vs_position_error_fit} shows an \SI{83}{\percent} lower mean position error over the similarly-setup 'Unaware' approach.
At a mean velocity \SI{28}{\percent} slower than 'Unaware' approach (see Fig. \ref{fig:soc_vs_velocity_fit}), there is also a significant but expected reduction in the peak position error as shown in Figure \ref{fig:soc_vs_peak_rmse}.
However, even with the conservative trajectory plan, the controller exceeded the planned thrust budget, in order to overcome the disturbances in flight, as signified by the peaks above the 'Available Thrust' curve in Figure \ref{fig:conservative_thrust}.

In the comparison shown in Figure \ref{fig:soc_vs_performance}, our proposed `Replanned` approach clearly showcases its advantages by bringing in a reduction of \SI{30}{\percent} in the mean position error while flying \SI{15}{\percent} faster than the non time-optimal `Conservative` approach.
Both the `Conservative` and `Replanned` approaches attained their peak position error at the start of the trajectory but `Replanned` approach consistently maintained a lower mean position error from thereon.
Even while flying faster, the improvement in the mean position error for 'Replanned` approach can be directly attributed to the restrained use of collective maximum thrust as shown in Figure \ref{fig:replanned_thrust} as opposed to the multiple limit breaches by the `Conservative` approach as shown in Figure \ref{fig:conservative_thrust}.

In conclusion, through this ablation study, we firmly establish that time optimality can be preserved in the face of shrinking propulsion limits of the \ac{uav}, even in highly demanding flight scenarios.
Our novel approach demonstrates clear advantages of thrust-awareness in an \ac{nmpc} formulation both with and without online replanning.
}

\begin{figure}
    \centering
    \begin{subfigure}[b]{\linewidth}
        \centering
        \includegraphics[width=\linewidth]{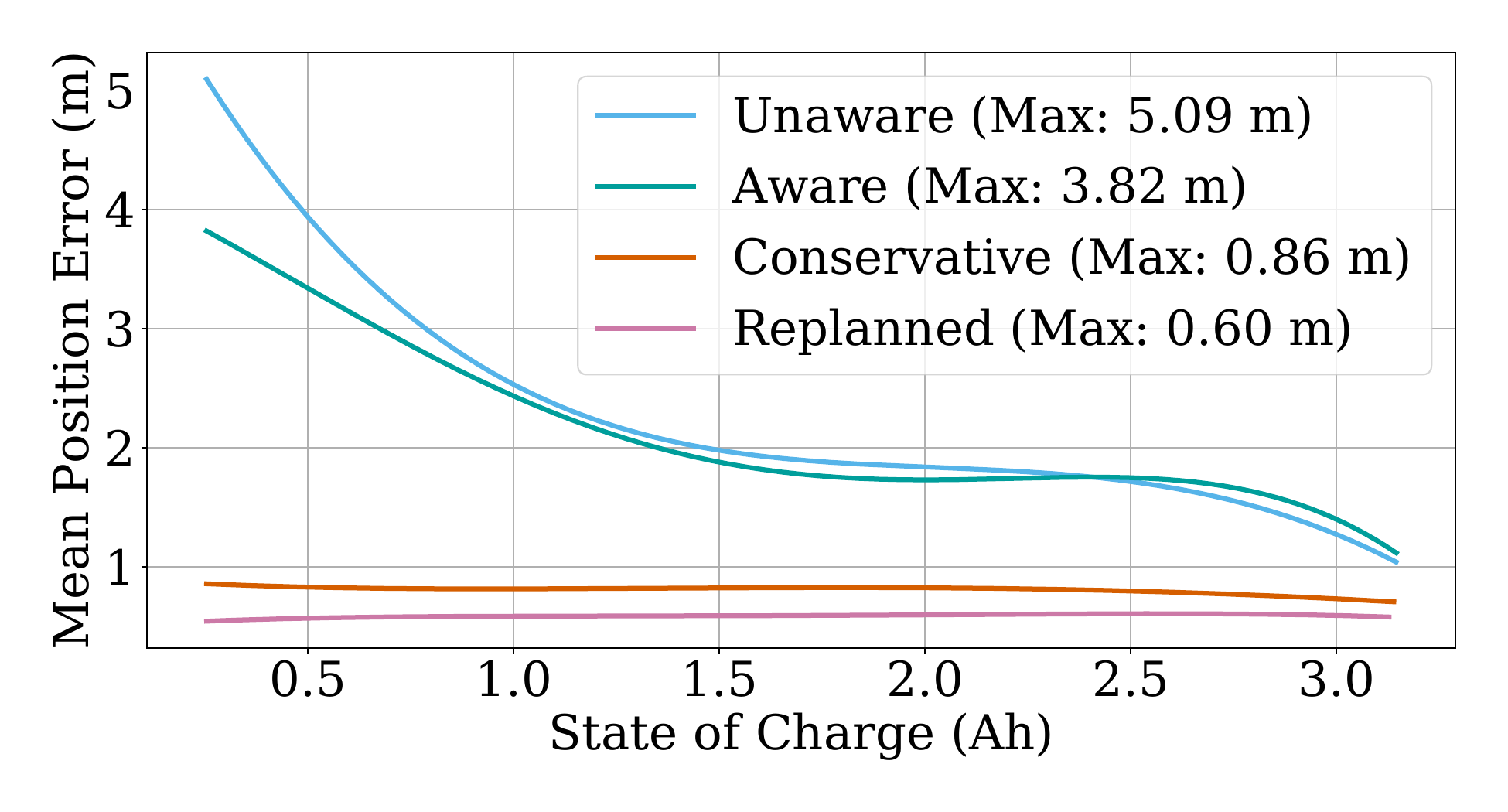}
        \caption{Mean position error throughout the battery range.}
        \label{fig:soc_vs_position_error_fit}
    \end{subfigure}
    \begin{subfigure}[b]{\linewidth}
        \centering
        \includegraphics[width=\linewidth]{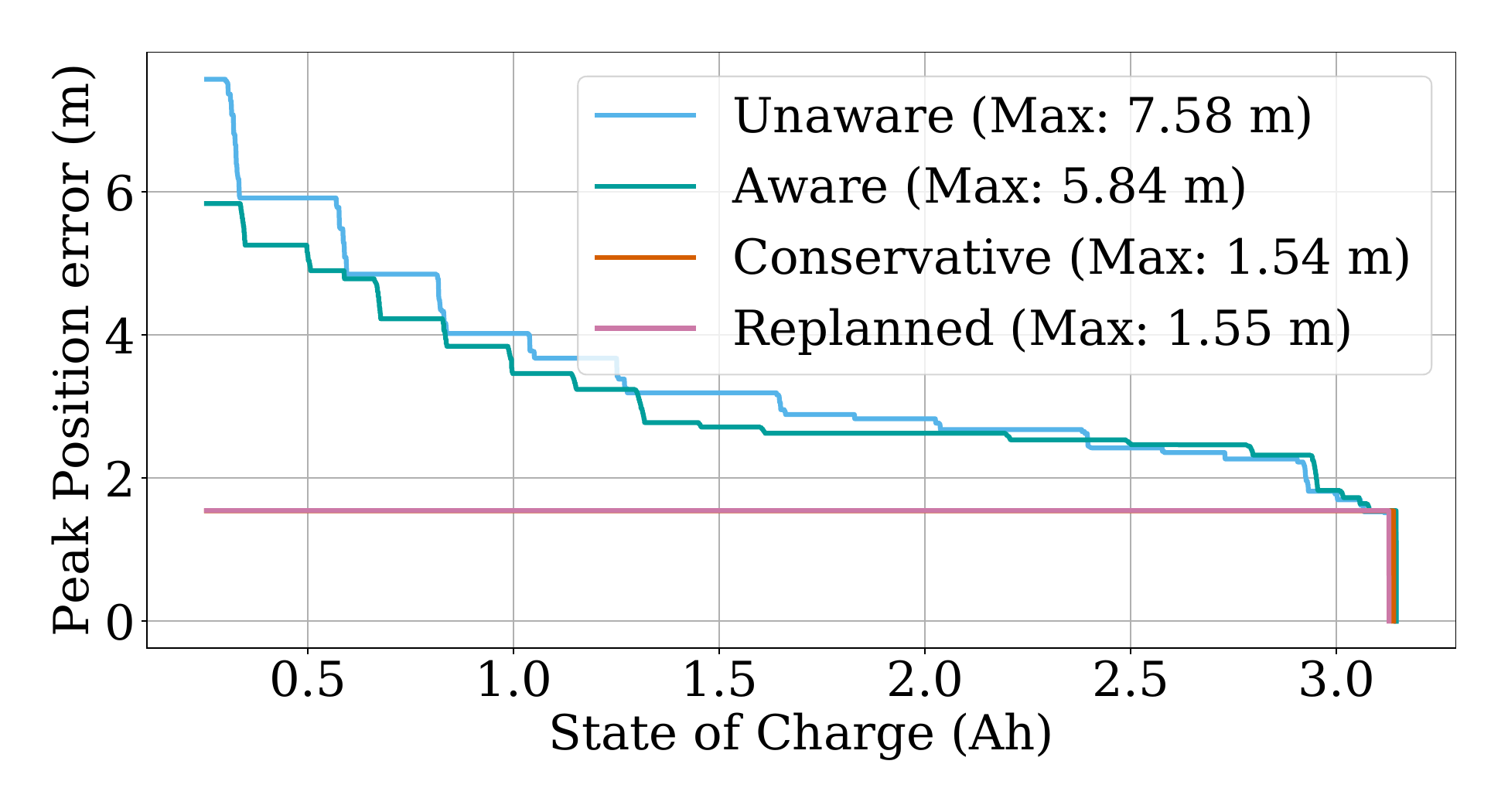}
        \caption{Peak position error throughout the battery range.}
        \label{fig:soc_vs_peak_rmse}
    \end{subfigure}
    \begin{subfigure}[b]{\linewidth}
        \centering
        \includegraphics[width=\linewidth]{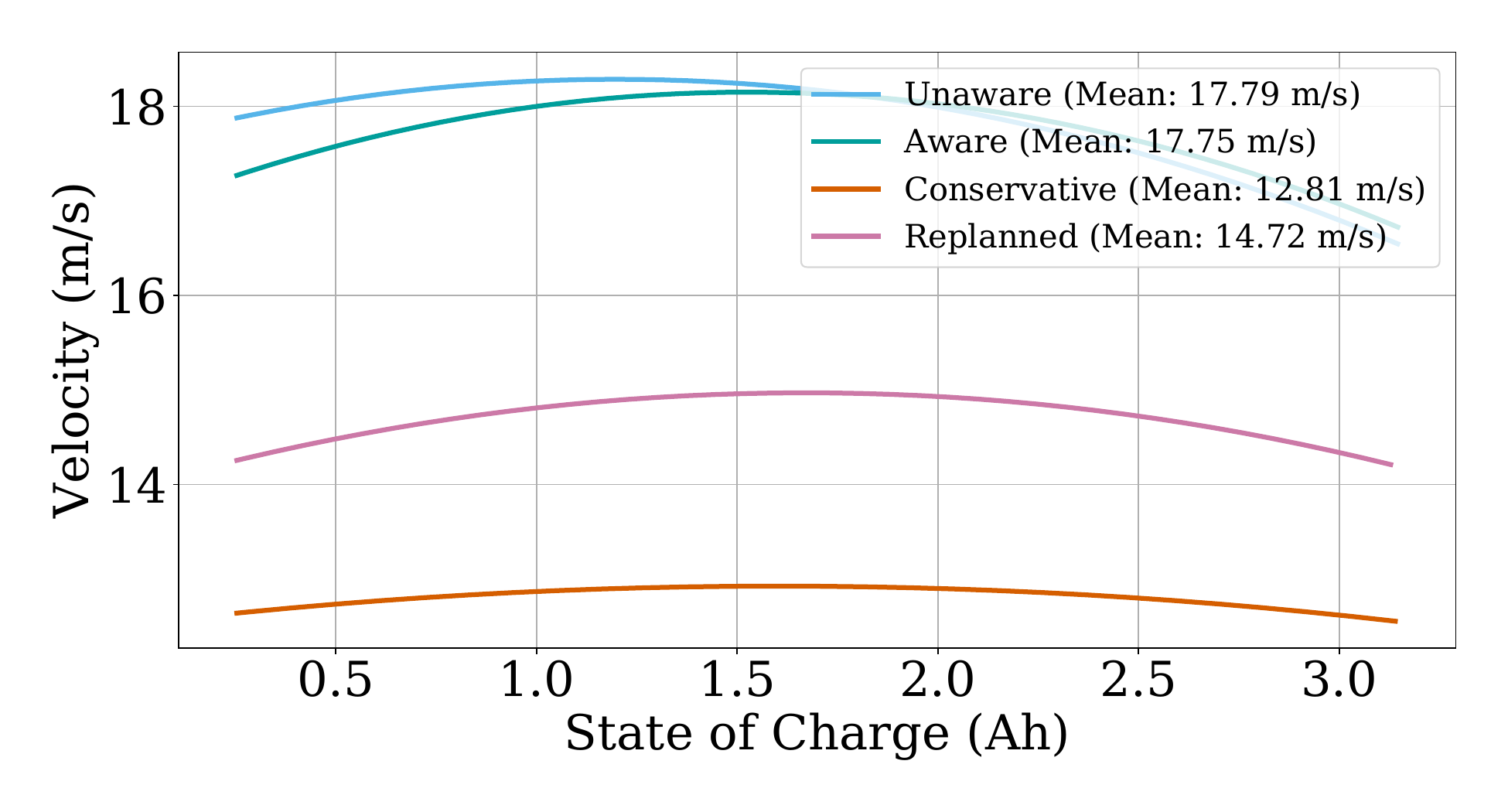}
        \caption{Mean velocity throughout the battery range.}
        \label{fig:soc_vs_velocity_fit}
    \end{subfigure}
    \caption{Comparison of position and velocity performance metrics for all four approaches through the battery range.}
    \label{fig:soc_vs_performance}
\end{figure}

\section{Conclusion}
\label{sec:con}

In this paper, we have presented a novel multi-variate polynomial model for the prediction of electro-mechanical characteristics of a motor-propeller-battery system to achieve real-time prediction of the voltage, current, power, and maximum available thrust of a quadrotor platform. 
The proposed model was integrated into an \ac{nmpc} framework to account for the dynamic variations in the maximum available thrust during the battery discharge cycle. 
This model was combined with a replanning approach to achieve improved trajectory tracking performance during an agile high-speed flight.
The proposed modelling was tested and studied in real-world experiments, while the real-time replanning approach was demonstrated in simulation experiments.
It was shown that even without replanning, our thrust-aware controller outperformed an unaware controller by \SI{25}{\percent} in mean position error, and by adding trajectory replanning, time-optimality could be conserved while reducing the mean position error by \SI{88}{\percent}.






\if\IEEE1
    \bibliographystyle{IEEEtran}
\else
    \bibliographystyle{cas-model2-names}
\fi
\bibliography{references}
%
\end{document}